\documentclass{article}

\usepackage[eandd, preprint, nonatbib]{neurips_2026}

\usepackage[utf8]{inputenc} %
\usepackage[T1]{fontenc}    %
\usepackage{hyperref}       %
\usepackage{url}            %
\usepackage{booktabs}       %
\usepackage{amsmath}        %
\usepackage{amssymb}
\usepackage{amsfonts}       %
\usepackage{nicefrac}       %
\usepackage{microtype}      %
\usepackage{xcolor}         %

\usepackage{cite}
\usepackage{enumitem}
\usepackage{graphicx}
\usepackage{tikz}
\usetikzlibrary{positioning}
\usepackage{subcaption}
\usepackage{wrapfig}       %
\usepackage{multirow}
\usepackage{montserrat}
\usepackage[scaled=0.85]{FiraMono}

\usepackage{float}         %
\usepackage{minted}
\setminted{
	fontsize=\scriptsize,
	breaklines,
}
\usepackage{tcolorbox}
\tcbuselibrary{minted,skins}
\definecolor{codebg}{HTML}{F7F7F8}

\newenvironment{denseitemize}{
	\begin{itemize}[topsep=2pt, partopsep=0pt, leftmargin=1.5em]
		\setlength{\parskip}{0pt}
		\setlength{\parsep}{0pt}
	}{\end{itemize}}

\renewcommand{\thefootnote}{\fnsymbol{footnote}}

\title{OpenG2G: A Simulation Platform for\\AI Datacenter--Grid Runtime Coordination}

\author{%
  Jae-Won Chung\thanks{Equal contribution.}\quad Zhirui Liang\footnotemark[1]\quad Yanyong Mao \\
  \textbf{Jiasi Chen} \quad \textbf{Mosharaf Chowdhury} \quad \textbf{Vladimir Dvorkin} \\
  \\
  University of Michigan  %
}

\begin{document}

\maketitle
\renewcommand{\thefootnote}{\arabic{footnote}}
\setcounter{footnote}{0}

\begin{abstract}
AI's growing compute demand and new datacenter buildouts present major capacity and reliability challenges for the electricity grid, leading to multi-year interconnection delays for new datacenters and bottlenecking AI growth.
To ease this strain, datacenters increasingly offer rapid power flexibility in response to grid signals, where the datacenter can increase or decrease its power consumption by adapting its workload in real time.

In order to understand the impact of large datacenters on the grid and to facilitate the design of effective coordination strategies, we build \emph{OpenG2G}, a simulation platform for \emph{AI datacenter--grid runtime coordination}.
We show that OpenG2G is capable of answering a wide range of coordination questions by allowing users to implement and compare various control paradigms (including classic, optimization, and learning-based controllers), and quantify how AI model and deployment choices affect datacenter flexibility and coordination outcomes.
This versatility is enabled by OpenG2G's modular and extensible architecture: a datacenter backend driven by real measurements of production-grade AI services, a grid backend built on high-fidelity grid simulators, and a generic controller interface that closes the loop between them.
We describe the design of OpenG2G and demonstrate its usefulness through realistic grid scenarios and AI workloads.\footnote{OpenG2G is open-sourced at \url{https://github.com/gpu2grid/openg2g}.}

\end{abstract}

\section{Introduction}\label{sec:introduction}

AI workloads consume large and rapidly growing amounts of energy~\cite{LBNL2024DataCenterEnergy,IEADataCenterCooling2025,bloombergnef25,cbre2025}.
A single modern AI datacenter easily draws tens of megawatts, and many planned ones are sized in gigawatts~\cite{openai-stargate,xai-colossus,meta-hyperion,google-nextera}.
For scale, 1 GW is roughly $1.7\times$ the average power draw of San Francisco~\cite{ca-electricity-data}.

The scale of these loads introduces challenges for grid operation.
For instance, grids operate under tight voltage limits that protect equipment and prevent cascading service interruptions~\cite{PGE_Rule2}.
Large loads like AI datacenters can introduce significant disruptions to keeping such constraints satisfied in the existing grid.
Therefore, grid operators may need to build new generation capacity, expand transmission and distribution infrastructure to deliver power reliably, maintain additional reserve margins to manage datacenter variability, and mitigate the risk of service interruptions when such loads are not properly accommodated.
This is one of the leading causes of multi-year delays in datacenter buildouts, which is bottlenecking AI capacity expansion and AI progress in general~\cite{ai-grid-impact-arxiv25}.

To help address this challenge, datacenters are increasingly willing to offer \emph{power flexibility}~\cite{google-flexibility-blog25,google-flexibility-blog26,colangelo2025ai} to bring more capacity online faster.
For instance, when the grid is experiencing larger load from other sources, the datacenter can reduce its power draw; conversely, when the grid has excess power,\footnote{This can arise from imperfect predictions of renewable generation and electricity demand.} the datacenter can increase its load, subject to its capacity limit, to absorb the surplus.
This is made possible by the wide range of control knobs that AI workloads in the datacenter admit, such as the choice of models, deployment configurations, and runtime controls, which can be adjusted to achieve tradeoffs between the datacenter's service performance and power draw without halting service~\cite{mlenergy-benchmark-neurips25,mlenergy-benchmark-v3-arxiv26}.
We refer to this runtime control problem of reasoning over datacenter knobs, grid operating decisions, and shared constraints as \emph{AI datacenter--grid runtime coordination} (\S\ref{sec:background}).\footnote{We note that our scope is runtime control, not planning, siting, or grid interconnection.}

\looseness=-1
However, this coordination problem is not well understood yet, including how to design effective controllers and how AI model and deployment choices affect datacenter flexibility.
This is because existing research is fragmented across the datacenter and grid sides, varying in realism and modeling assumptions, and lacks a unifying simulation framework on which to build (\S\ref{sec:related}).
To that end, we built \emph{OpenG2G}, an open-source library for simulating AI datacenter--grid runtime coordination (\S\ref{sec:design}).
OpenG2G composes three pluggable components around a simulation loop: a datacenter backend driven by measurements from production-grade AI services (e.g., \cite{mlenergy-benchmark-neurips25,mlenergy-benchmark-v3-arxiv26}), a grid backend that wraps around traditional grid simulators (e.g., \cite{opendss,opendssdirect}), and a generic controller implementation interface that exposes datacenter and grid state and emits control actions that act on either side.
Swapping in a new AI workload, grid topology/simulator, or controller amounts to writing a Python subclass.
The controller interface is flexible enough to support various simple to complex control paradigms, allowing head-to-head comparisons on identical scenarios.

We demonstrate the usefulness of OpenG2G from two perspectives using inference as a critical AI workload.\footnote{Inference is known to account for 80--90\% of AI compute demand~\cite{nvidia-inference-estimation,aws-inference-estimation,patterson2021carbon,polca-asplos24}.}
First, we use the controller interface to implement classical feedback controllers (e.g., online feedback optimization (OFO) \cite{bernstein2019real,hauswirth2024optimization}, droop control \cite{bollen2005voltage}) and learning-based controllers (e.g., Proximal Policy Optimization (PPO)~\cite{ppo-arxiv17}) and compare their coordination outcomes head-to-head, showing how the controller design space can be explored with OpenG2G (\S\ref{sec:controllers}).
Second, we vary AI model and deployment choices and quantify how each shapes the datacenter's \emph{feasible power range} (MW), and through it, the room available for runtime coordination (\S\ref{sec:ml-impact}).

In summary, we make the following contributions:
\begin{denseitemize}
  \item We build OpenG2G, an open-source library for simulating AI datacenter--grid runtime coordination, whose datacenter, grid, and controller components bridge systems, machine learning, and grid engineering communities to address the energy/grid challenge of modern AI electricity demand.
  
  \item OpenG2G captures metrics spanning both AI datacenter (e.g., throughput, latency) and power systems (e.g., grid voltages), enabling standardized comparison and evaluation of the impact of various control strategies (e.g., droop, OFO, PPO) and AI model and deployment choices on datacenter--grid coordination outcomes.
  
  \item We simulate coordination between modern LLM (e.g., Llama, Qwen, GPT-OSS) inference and grid operations and reveal favorable trade-offs between AI and grid operational metrics, demonstrating OpenG2G's potential to inform actionable design decisions for AI datacenter projects.
\end{denseitemize}

\section{Background and Problem Formulation}\label{sec:background}

\subsection{AI Datacenters and the Grid}\label{sec:background-dcgrid}

\begin{figure}[t]
\centering
\includegraphics[width=0.83\linewidth]{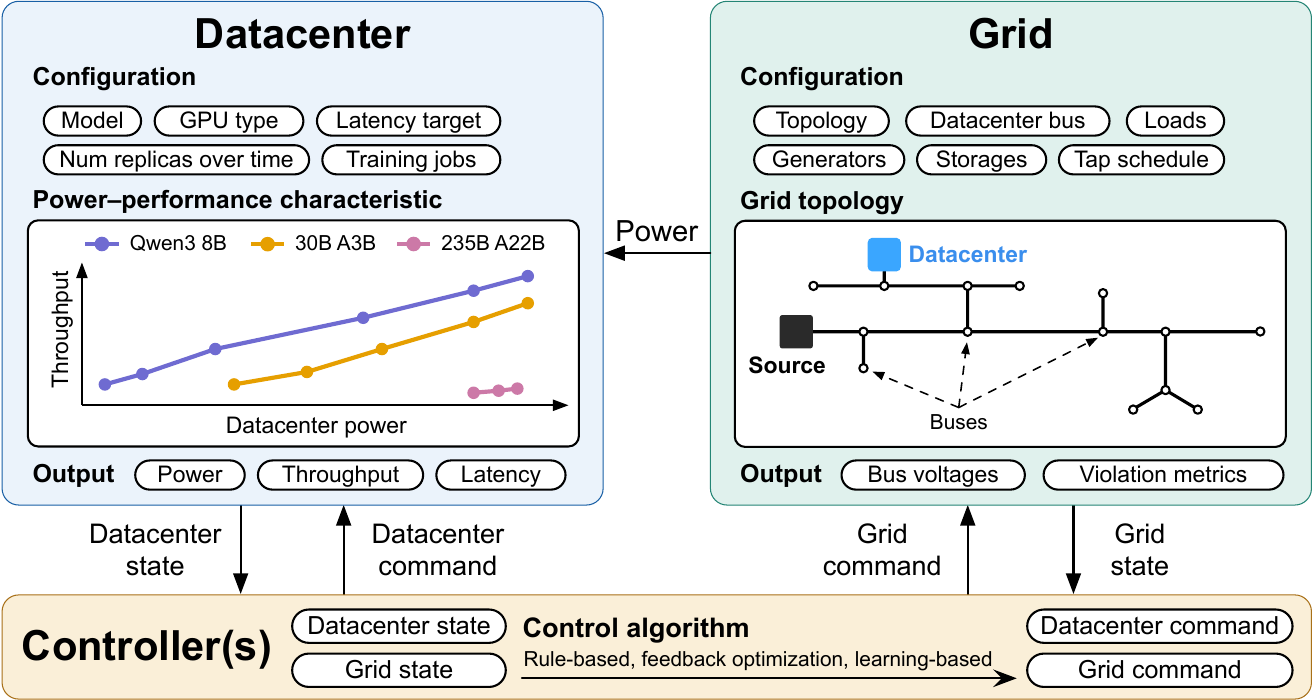}
\caption{
  Overview of OpenG2G's architecture.
  OpenG2G composes three pluggable components around a simulation loop that welds them together and emits datacenter and grid metrics.
}
\label{fig:overview}
\end{figure}

Each power consumer, including datacenters, attaches to the grid at a \emph{bus}, with buses connected by power lines.
Bulk power enters the modeled network at a \emph{source} bus: a substation in distribution settings, or a generator or higher-voltage interconnection in transmission settings.
Figure~\ref{fig:overview}'s Grid box illustrates a simple distribution grid with one datacenter load connected.

The grid's electrical state must stay within rated ranges (e.g., bus voltages within a narrow band around the nominal value, line currents below their thermal limits) to avoid equipment damage and service interruptions.
Operators maintain these constraints with grid-side devices such as on-load tap changers, capacitor banks, and energy storage, whose actuation latencies span from sub-second for power-electronic devices to minute-scale for mechanical adjustments.

An AI datacenter attaches as a single large load whose power draw is shaped by the workload it runs.
For inference, model choice (size and architecture), batch size, precision, hardware (GPU type), parallelism, and application constraints collectively determine latency, throughput, and power~\cite{mlenergy-benchmark-neurips25,mlenergy-benchmark-v3-arxiv26}.

\subsection{Coordination Problem}\label{sec:background-coordination}

From the datacenter operator's perspective, permitting degraded performance from time to time is better than having no service at all while waiting for grid interconnection.
Thus, to unblock infrastructure-bound capacity expansion, AI datacenter operators are increasingly committing to providing datacenter power flexibility to the grid~\cite{google-flexibility-blog25,google-flexibility-blog26}; when the grid is stressed, the datacenter can be signaled to reduce its power draw, and when the grid is in excess supply, the datacenter can in fact increase its power draw to help stabilize the grid.
This is feasible because AI workloads admit a wide range of control knobs (e.g., batch size; \S\ref{sec:background-dcgrid}) that can be rapidly adjusted to shape datacenter power without halting service.
Realizing this flexibility well requires runtime control that reasons over datacenter-side and grid-side state and dispatches actions to either side.

We refer to this problem as \emph{AI datacenter--grid runtime coordination}: AI datacenters and the grid form a closed-loop system through their shared power trajectory, with controllers on either side acting at millisecond-to-minute timescales.
Each side has its own goals, objectives, and runtime knobs, so a one-sided policy that improves AI throughput can shrink the grid's electrical flexibility, and a grid-stabilizing action can consume AI service capacity or violate AI service targets.
Studying this rigorously requires a simulation framework that combines realistic AI workload behavior with grid simulation behind a common control interface, which no existing tool provides.

\section{OpenG2G}\label{sec:design}

OpenG2G is organized around three pluggable components (datacenter, grid, and controller) and a generic, multi-rate simulation loop that drives them (Figure~\ref{fig:overview}).
To assemble a scenario, the user instantiates each component with its inputs, defines their connections, and hands them to the simulation loop.
Each simulation tick, each component is advanced and produces metrics.

\paragraph{Datacenter.} \looseness=-1
The datacenter component simulates one or more AI clusters, with configurable inputs (model, GPU, latency target, replica schedule, training jobs) and per-step outputs of power, throughput, and latency (Figure~\ref{fig:overview}, Datacenter box).
Controllers act on it by issuing commands such as changing batch size and scaling the number of replicas.
The default backend replays power, throughput, and latency measurements from the ML.ENERGY Benchmark dataset~\cite{mlenergy-benchmark-neurips25,mlenergy-benchmark-v3-arxiv26}, behaving like a real datacenter without requiring costly large-scale deployments.
Scheduled training jobs can also be overlaid on inference workloads, with power traces distilled from real large-model training measurements~\cite{kareus-arxiv26}.

\paragraph{Grid.}
The grid component simulates the electricity grid, taking configurable inputs such as topology, tap-position trajectories, loads, generators, and energy storage, and reporting per-step bus voltages and violation metrics (Figure~\ref{fig:overview}, Grid box).
Controllers interact with the grid by issuing commands such as changing regulator tap positions.
The default implementation models a distribution feeder whose primary control objective is voltage regulation: demand reduction helps mitigate undervoltage, while demand increase helps mitigate overvoltage.
This module wraps around the widely used OpenDSS~\cite{opendss,opendssdirect} simulator and includes support for standard IEEE test feeders~\cite{ieee-test-feeders}.
Users can add fixed or time-varying loads and generators, attach energy storage, or provide a custom grid definition.

\paragraph{Controller.}
The controller is the closed-loop policy that coordinates the datacenter and grid by reading the current datacenter and grid state and sending a list of datacenter and/or grid commands (Figure~\ref{fig:overview}, Controller box).
Any controller is a subclass of the abstract \texttt{Controller} base class, which defines the minimal interface for the simulation loop to interact with it.
This modular structure allows users to experiment with different controllers on the same datacenter and grid scenario, and to test the same controller across different datacenter and grid scenarios.
When multiple datacenters share a grid, control can be centralized (one controller per grid) or decentralized (one controller per datacenter).

\paragraph{Simulation flow.} \looseness=-1
At each tick, the simulation loop advances the datacenter, grid, and controller at their own native fractional rates (e.g., one tick every 0.1 seconds), brokers the observation--command exchange between them, and logs the metrics and events each component emits.
Multi-rate scheduling matters because each component may have different natural cadences: grid simulators may be too expensive to run frequently, whereas one may wish to capture the datacenter's full power timeline at the finest granularity allowed by the underlying dataset.
Logging, seeding, and event bookkeeping are centralized in the loop, so every scenario is reproducible and allows post-hoc analysis through a consistent interface and data format.

\paragraph{Extensibility.}
Users can extend OpenG2G by subclassing the abstract base of any pluggable component (datacenter, grid, or controller) and implementing a list of abstract methods.
To add a new AI workload, users add a measurement record to the data pipeline, supply their own traces, or substitute a custom \texttt{DatacenterBackend} subclass; to add a new grid simulator or controller, they implement a \texttt{GridBackend} or \texttt{Controller} subclass, respectively (Appendix~\ref{sec:appendix-extensibility} provides code examples and explanations).
Because the simulation loop only orchestrates tick scheduling and the exchange of observations and commands, it is agnostic to internal implementations of each component; users do not need to modify the loop to add new datacenter backends, grid backends, or controllers.
The next two sections (\S\ref{sec:controllers} and \S\ref{sec:ml-impact}) demonstrate this extensibility and its usefulness.

\section{Evaluating Controllers for AI Datacenter--Grid Coordination}\label{sec:controllers}

This section demonstrates how OpenG2G can be used to implement and evaluate different control paradigms for AI datacenter--grid runtime coordination.
We use inference \emph{batch size} as the primary datacenter-side knob for voltage regulation.
Batch size is well suited for runtime coordination because it can be changed rapidly without redeploying inference servers or halting service.
Moreover, power, throughput, and inter-token latency are all sensitive to batch size~\cite{mlenergy-benchmark-neurips25,mlenergy-benchmark-v3-arxiv26}, so this single knob exposes a meaningful operating range for balancing grid support and inference performance.\footnote{OpenG2G supports other control knobs like changing the number of server replicas and is extensible to more.}

The basic control principle is simple: increase batch size during overvoltage to soak up power, and decrease it during undervoltage to shed load.
The challenge is choosing the right adjustment magnitude, which should be large enough to mitigate voltage violations, but not so large that it unnecessarily reduces throughput or violates latency targets.
This requires either explicit models of how batch size affects power, throughput, and latency, or policies learned through grid interactions. For simplicity, replicas of the same LLM within a datacenter share one batch size, while batch sizes may differ across LLMs and datacenters.

\paragraph{Controllers.}
We compare three controllers, all acting on inference batch size: a classic \emph{droop} controller, an online feedback optimization (\emph{OFO}) controller~\cite{g2g-powerup26}, and a learning-based proximal policy optimization (\emph{PPO}) controller~\cite{ppo-arxiv17}.
Droop reacts linearly to the worst voltage violation without explicitly modeling throughput or latency.
OFO uses fitted batch-to-power/throughput/latency functions and grid sensitivity matrices that maps datacenter power changes to bus-voltage changes.
PPO is a model-free actor--critic controller that learns from grid interaction, observing voltages and LLM states and outputting per-LLM actions to decrement, hold, or increment batch size under a multi-objective reward.
For multi-datacenter systems, PPO is centralized, while droop and OFO run per datacenter with shared parameters and control rules.
The \emph{No Coordination} baseline fixes all batch sizes at 128, a value supported by all models with reasonable latency.
Appendix~\ref{sec:appendix-controller-math} provides full formulations and hyperparameters.
  
\paragraph{Setup.}
We evaluate the controllers on the standard IEEE 13-, 34-, and 123-bus feeders~\cite{ieee-test-feeders} hosting one, two, and four datacenters, respectively.
Appendix~\ref{sec:appendix-grid} provides visualizations of the feeders and details of the datacenter placement, generators, and loads.
Datacenters serve Llama 3.1 8B, 70B, and 405B~\cite{llama3-arxiv24}, and Qwen 3 30B A3B and 235B A22B~\cite{qwen3-arxiv25}.
Each scenario is a randomly-sampled combination of generation and load profiles, inference replica-count ramps, and an optional coincident LLM training overlay, run for $3{,}600$ seconds with a 1~s control step; Appendix~\ref{sec:appendix-ppo-pipeline} details the scenario library.
We evaluate each (controller, feeder) pair five times across 50 disjoint test scenarios and report the mean and standard deviation across the five runs.
When the results are deterministic, standard deviation is omitted.
  
\paragraph{Metrics.}
The main coordination outcome metric is the \emph{integral voltage violation} (pu$\cdot$s), which accumulates the total bus-phase voltage deviation outside the allowable band ($\pm$ 5\%) over the episode and captures both duration and severity of voltage excursions.
We also report \emph{token throughput} (tok/s), the mean token-output rate across all deployed LLMs. %
Mathematical definitions are in Appendix~\ref{sec:appendix-setup}.

\subsection{Controller Evaluation in Terms of Voltage Regulation}\label{sec:controllers-eval}

\begin{figure}[t]
  \centering
  \includegraphics[width=0.68\linewidth]{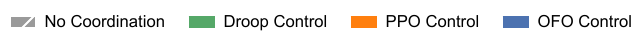}
  \begin{minipage}[t]{0.32\linewidth}\centering
    \subfloat[IEEE 13-bus]{\label{fig:eval_voltage_a}\includegraphics[width=\linewidth]{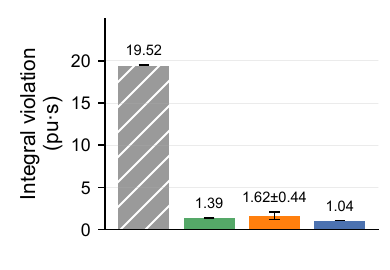}}
  \end{minipage}\hspace{0.5em}%
  \begin{minipage}[t]{0.32\linewidth}\centering
    \subfloat[IEEE 34-bus]{\label{fig:eval_voltage_b}\includegraphics[width=\linewidth]{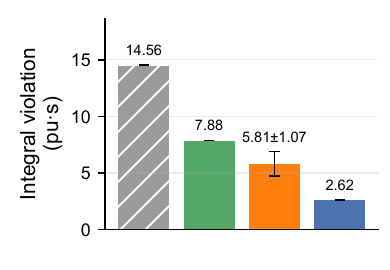}}
  \end{minipage}\hspace{0.5em}%
  \begin{minipage}[t]{0.32\linewidth}\centering
    \subfloat[IEEE 123-bus]{\label{fig:eval_voltage_c}\includegraphics[width=\linewidth]{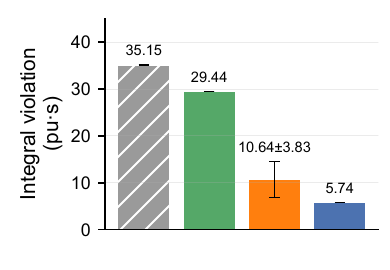}}
  \end{minipage}
  \caption{Controllers' voltage regulation performance diverges with feeder complexity. Bars show the mean integral voltage violation over 50 evaluation scenarios. For PPO, we train five policies with different random seeds and report the mean $\pm$ standard deviation across the five trained controllers.}
  \label{fig:cross_system_voltage}
  \vspace{-1.0em}
\end{figure}

Figure~\ref{fig:cross_system_voltage} compares the mean integral voltage violation using 50 evaluation scenarios across three distribution feeders.
We can observe the trend of how controller performance changes as the grid-control problem becomes more complex.
On the single-datacenter IEEE 13-bus feeder, all three controllers substantially reduce voltage violations relative to No Coordination, suggesting that even simple feedback can be effective when the system is small and control is spatially concentrated.
As the feeders become larger and multiple datacenters must coordinate across a more complex topology, the performance gap among the three controllers widens.
Thus, OFO serves as an information-rich performance reference, while droop and PPO show how far simpler or model-free controllers can go under the same OpenG2G evaluation framework.

\subsection{Trade-off Between Voltage Regulation and Throughput}

\begin{figure}[t]
    \centering
    \includegraphics[width=0.74\linewidth]{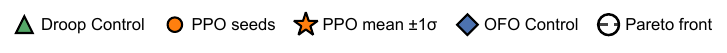}
    \begin{minipage}[t]{0.32\linewidth}\centering
      \subfloat[IEEE 13-bus]{\label{fig:tradeoff_a}\includegraphics[width=\linewidth]{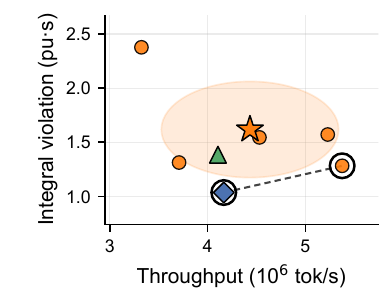}}
    \end{minipage}\hspace{0.5em}%
    \begin{minipage}[t]{0.32\linewidth}\centering
      \subfloat[IEEE 34-bus]{\label{fig:tradeoff_b}\includegraphics[width=\linewidth]{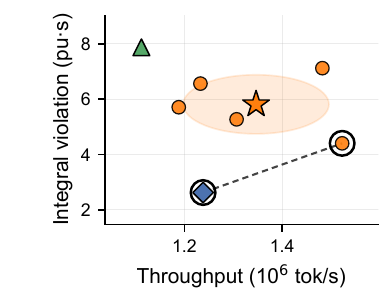}}
    \end{minipage}\hspace{0.5em}%
    \begin{minipage}[t]{0.32\linewidth}\centering
      \subfloat[IEEE 123-bus]{\label{fig:tradeoff_c}\includegraphics[width=\linewidth]{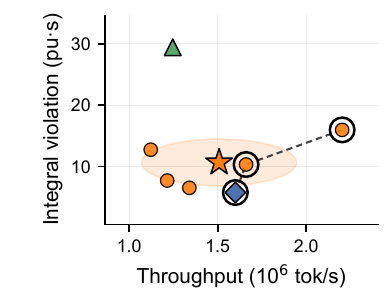}}
    \end{minipage}
    \caption{Voltage regulation and throughput tradeoffs of controllers. Orange dots show the five PPO seeds, while the orange star and shaded ellipse indicate the seed mean and $\pm 1\sigma$ spread. OFO remains Pareto-optimal across all systems.}
    \label{fig:tradeoff}
  \end{figure}

Figure~\ref{fig:tradeoff} shows that voltage regulation is not free: controllers that more aggressively reduce voltage violations may also reduce batch sizes and thus throughput.
PPO achieves higher mean throughput than OFO on the IEEE 13- and 34-bus feeders by accepting larger integral violations, while OFO remains Pareto-optimal in the throughput--violation plane across all three systems.
This difference reflects how each controller weighs competing objectives in its objective or reward function, rather than an inherent ranking between PPO and OFO.\@
Because the controller is operated by the datacenter owner, these weights provide a practical lever for trading throughput against voltage support based on the economic value of the running workload.
For example, the datacenter owner could enter into an agreement with the grid operator that specifies compensation for power flexibility, and then tune the objective weights to provide the economically appropriate level of grid support.
Droop control, in contrast, lands at neither extreme of the trade-off in these experiments.
Appendix~\ref{sec:appendix-controller-perf} reports the full per-controller, per-feeder evaluation results and parameter sweeps for OFO and PPO, further illustrating the economic and operational trade-offs induced by different control objectives.

\subsection{Deep Dive into Controller Behavior}

\begin{figure}[t]
    \centering
    \includegraphics[width=0.68\linewidth]{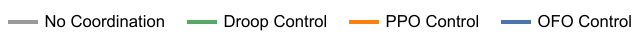}

    \begin{minipage}[t]{0.30\linewidth}\centering
      \subfloat[Llama 3.1 8B]{\label{fig:batch_a}\includegraphics[width=\linewidth]{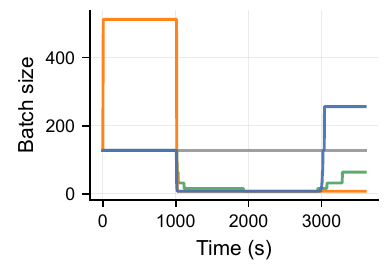}}
    \end{minipage}\hspace{0.5em}%
    \begin{minipage}[t]{0.30\linewidth}\centering
      \subfloat[Llama 3.1 405B]{\label{fig:batch_b}\includegraphics[width=\linewidth]{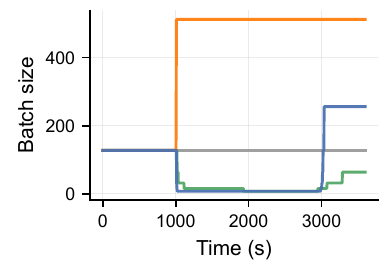}}
    \end{minipage}\hspace{0.5em}%
    \begin{minipage}[t]{0.30\linewidth}\centering
      \subfloat[Qwen 3 235B A22B]{\label{fig:batch_c}\includegraphics[width=\linewidth]{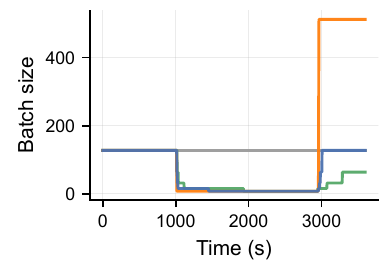}}
    \end{minipage}
    \caption{Per-model batch-size responses under the same voltage-disturbance scenario. The datacenter deploys five LLMs; we show three representative models, with the other two exhibiting similar trends. Droop, PPO, and OFO achieve voltage regulation through distinct per-model batch-size policies.}
    \label{fig:batch_size}
  \end{figure}

How do different controllers utilize the batch size control knob for voltage regulation?
Figure~\ref{fig:batch_size} compares the batch size trajectories of droop, PPO, and OFO under the same voltage-disturbance scenario on the IEEE 13-bus feeder setup. 
All three controllers substantially improve over the No Coordination baseline (\S\ref{sec:controllers-eval}); the main difference is in how they act to achieve these gains.
Droop and OFO produce qualitatively similar step-based responses to voltage error: both reduce batch size around $1000$ s to shed load during undervoltage and increase batch size around $3000$ s to restore load during overvoltage.
However, droop uses a fixed response gain and only targets voltage violations, whereas OFO adapts its effective step through the voltage-constraint dual variable and jointly balances throughput, voltage regulation, and latency violation.
Consequently, OFO generally maintains larger batch sizes than droop (for example, rising rapidly around 3000 s) while achieving the smallest violation.
PPO instead learns a coordinated multi-model policy from data, allowing different models to take different corrective roles as long as the total reward is high.
This produces more diverse per-model trajectories while preserving the same high-level behavior: reducing load during undervoltage and increasing load during overvoltage.

\emph{Takeaway.}
Droop, PPO, and OFO controllers all reduce voltage violations relative to No Coordination, but they trace different paths through the throughput--violation plane.
As feeder scale grows, the gap widens: information-rich OFO achieves the strongest voltage regulation, while model-free PPO retains higher throughput on smaller feeders at the cost of more violations.
Where each controller lands therefore depends on two factors: how much grid and datacenter information it has access to, and how it weighs throughput against voltage violation when optimizing performance.

\section{Evaluating the Impact of AI Choices on Feasible Power Range}\label{sec:ml-impact}

We use OpenG2G to investigate how AI model and deployment choices shape the datacenter's \emph{feasible power range} (MW), and through it, how the room available for runtime coordination changes.
For a fixed datacenter size, we vary AI model (\S\ref{sec:ml-impact-model-design}) and deployment (\S\ref{sec:ml-impact-deployment}) choices.

\paragraph{Setup.}
We adopt Section~\ref{sec:controllers}'s IEEE 13-bus, single-datacenter setup (Appendix~\ref{sec:appendix-setup}), but run a single model at a time.
Each model targets 50 ms per-token latency, except the larger Qwen 3 235B A22B~\cite{qwen3-arxiv25} which relaxes to 100 ms because 50 ms is infeasible on its 8-GPU deployment.
We run each scenario with No Coordination and OFO, and compare the integral voltage violation (pu$\cdot$s) between them to understand the room for runtime coordination; results are deterministic across five random seeds.
The number of LLM inference replicas is chosen so that the datacenter's peak power is 3 MW (e.g., 4,800 replicas for Qwen 3 8B on H100 GPUs) at the largest batch size meeting the latency target.
Anchoring all configurations to the same peak power allows any change in the feasible power range we observe across configurations to be attributable to the AI model and deployment choices under study, rather than to differences in deployed capacity.

\paragraph{Feasible power range.}
An intermediate metric we find to be important in gauging the power flexibility of the datacenter is its \emph{feasible power range} (MW), the span (max minus min) of feasible datacenter power consumption across LLM inference configurations that meet the token latency target.
Feasible power range corresponds to the horizontal span of the power--throughput curve (e.g., Figure~\ref{fig:model-size-b200-pareto}).
A larger feasible power range gives the controller more room to mitigate grid stress by adjusting the datacenter's power consumption.
When batch size is the primary control knob, the lower and upper endpoints of the feasible power range correspond to the smallest and largest batch sizes that meet the latency target, respectively.
As we will see, each AI model and deployment choice studied below lifts or lowers one of these endpoints, narrowing or widening the feasible power range.

\subsection{AI Model Choices}\label{sec:ml-impact-model-design}

\begin{figure}[t]
\centering
\includegraphics[width=\linewidth]{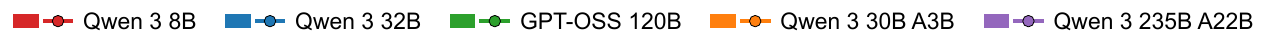}

\begin{minipage}[t]{0.36\linewidth}\centering
\subfloat[Integral voltage violation]{\label{fig:model-size-b200-violation}\includegraphics[width=0.9\linewidth]{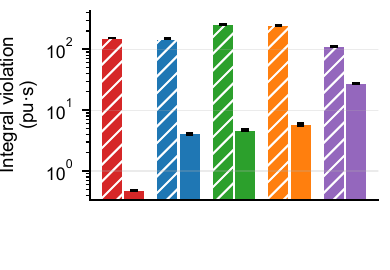}}
\end{minipage}\hspace{1.5em}%
\begin{minipage}[t]{0.36\linewidth}\centering
\subfloat[Power--throughput curve]{\label{fig:model-size-b200-pareto}\includegraphics[width=0.9\linewidth]{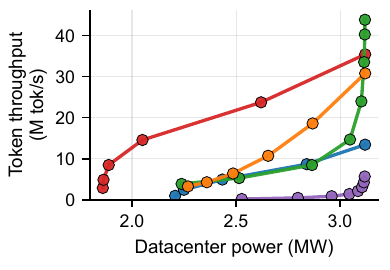}}
\end{minipage}
\caption{
Model size and architecture: five models served on B200 GPUs (1, 1, 1, 2, and 8 GPUs per replica respectively), ordered by decreasing feasible power range.
Hatched bars are No Coordination; solid bars use OFO.\@
Models with wider span in (b) show larger reductions in (a).
}\label{fig:model-size-b200}
\end{figure}

\paragraph{Model size and architecture.}
Figure~\ref{fig:model-size-b200} compares five models served on B200 GPUs, ordered left-to-right by decreasing feasible power range.
Qwen 3 8B (red) provides the largest feasible power range, allowing OFO to drive integral voltage violation well below 1 pu$\cdot$s; the other models have narrower ranges for structural reasons.
Qwen 3 32B (blue) has its feasible batch range cut short by the latency target, lowering the ceiling: the 50 ms target hits at batch size $128$ rather than $256$ of the 8B model.
Qwen 3 30B A3B (orange) is a mixture-of-experts (MoE) model: only $3$B parameters activate per token, but each forward pass routes tokens across many experts.
This keeps GPU activity higher than in a dense model of similar size even at small batch sizes, raising the power floor compared to the dense Qwen 3 8B model.
Qwen 3 235B A22B (purple) on 8 GPUs is large enough that even the smallest batch size leads to high GPU utilization and power draw, significantly raising the power floor.
Its token throughput is far lower than the other models because of its size.
Appendix~\ref{sec:appendix-h100-model-size} extends this study to H100, adding Llama 3.1 70B and 405B.

\begin{figure}[t]
\centering
\includegraphics[width=0.234\linewidth]{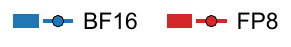}

\begin{minipage}[t]{0.36\linewidth}\centering
\subfloat[Integral voltage violation]{\label{fig:precision-violation}\includegraphics[width=0.9\linewidth]{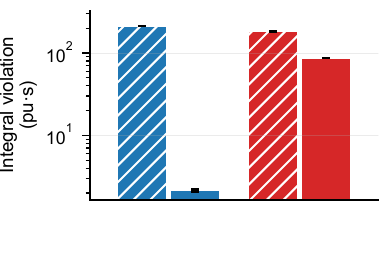}}
\end{minipage}\hspace{1.5em}%
\begin{minipage}[t]{0.36\linewidth}\centering
\subfloat[Power--throughput curve]{\label{fig:precision-pareto}\includegraphics[width=0.9\linewidth]{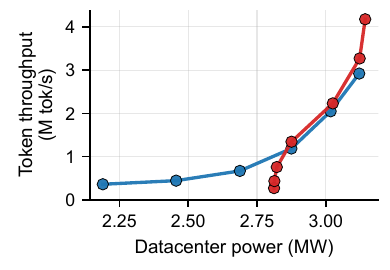}}
\end{minipage}
\caption{
Weight precision: Qwen 3 235B A22B on $8 \times$ H100 GPUs.
Hatched bars are No Coordination; solid bars use OFO.\@
BF16 has wider feasible power range; FP8 reaches higher throughput.
}\label{fig:precision}
\end{figure}

\paragraph{Weight precision.}
Figure~\ref{fig:precision} compares BF16 against FP8 for the Qwen 3 235B A22B model on $8 \times$ H100 GPUs.
At small batch sizes, LLM decoding has low arithmetic intensity, so the tensor cores are underutilized, and the GPUs draws well below peak power.
FP8 halves the bytes per weight, raising arithmetic intensity in this regime and lifting the power floor.
With OFO, this translates into a roughly $100\times$ drop in BF16's integral violation, but only a $2\times$ drop for FP8.
Appendix~\ref{sec:appendix-precision} provides two more BF16 and FP8 pairs that share the mechanism.

\subsection{AI Deployment Choices}\label{sec:ml-impact-deployment}

\begin{figure}[t]
\centering
\begin{minipage}[t]{0.36\linewidth}\centering
\includegraphics[width=0.6\linewidth]{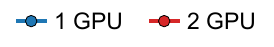}\\[-2pt]
\subfloat[GPT-OSS 120B, 1 vs 2 GPU]{\label{fig:parallelism-gpt-oss-b200}\includegraphics[width=0.9\linewidth]{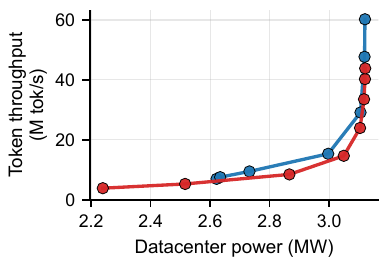}}
\end{minipage}\hspace{1.5em}%
\begin{minipage}[t]{0.36\linewidth}\centering
\includegraphics[width=0.6\linewidth]{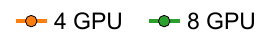}\\[-2pt]
\subfloat[Qwen 3 235B A22B, 4 vs 8 GPU]{\label{fig:parallelism-qwen-thinking-b200}\includegraphics[width=0.9\linewidth]{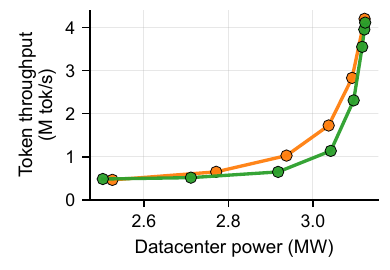}}
\end{minipage}
\caption{
Parallelism: (a) GPT-OSS 120B~\cite{gpt-oss-report} and (b) Qwen 3 235B A22B~\cite{qwen3-arxiv25}.\@
Doubling expert parallelism degree widens the feasible batch range in both pairs, but translates into a substantially wider feasible power range in (a) and barely any change in (b).
}\label{fig:parallelism-b200}
\end{figure}

\paragraph{Parallelism.}
Figure~\ref{fig:parallelism-b200} varies the expert parallelism degree for two models running on B200: GPT-OSS 120B~\cite{gpt-oss-report} at $1 \times$ vs $2 \times$ GPUs, and Qwen 3 235B A22B at $4 \times$ vs $8 \times$ GPUs.
Doubling parallelism for GPT-OSS 120B widens the feasible power range substantially.
This is because at small batch sizes, each GPU runs half the model shard, gets underutilized, and draws less power, which lowers the power floor.
On the other hand, doubling parallelism for Qwen 3 235B A22B does not show such an effect.
As the model is large, increasing parallelism and further sharding computation does not reduce per-GPU utilization even at small batch sizes, so the power floor does not drop.
Appendix~\ref{sec:appendix-h100-parallelism} provides a Qwen 3 30B A3B parallelism pair on H100.

\begin{figure}[t]
\centering
\includegraphics[width=0.344\linewidth]{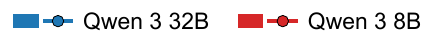}

\begin{minipage}[t]{0.36\linewidth}\centering
\subfloat[Integral voltage violation]{\label{fig:hardware-violation}\includegraphics[width=0.9\linewidth]{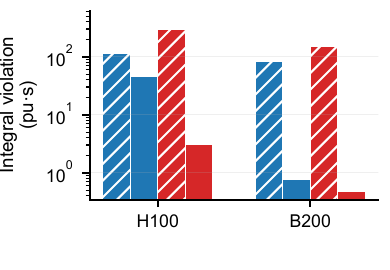}}
\end{minipage}\hspace{1.5em}%
\begin{minipage}[t]{0.36\linewidth}\centering
\subfloat[Power--throughput curve]{\label{fig:hardware-pareto}\includegraphics[width=0.9\linewidth]{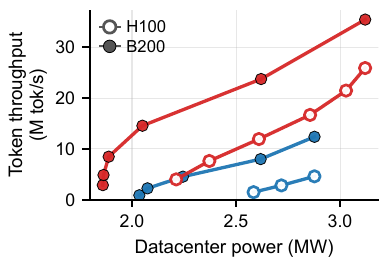}}
\end{minipage}
\caption{
GPU type (hardware generation): Qwen 3 8B and 32B~\cite{qwen3-arxiv25} on H100 vs B200.
In (a), hatched bars are No Coordination; solid bars use OFO.\@
In (b), open markers are H100 and filled markers are B200.
B200 provides wider feasible power range for both models.
}
\label{fig:hardware}
\vspace{-0.5em}
\end{figure}

\paragraph{GPU type.}
Figure~\ref{fig:hardware} shows integral voltage violation for two dense models on H100 (2023~\cite{h100-ga}) vs B200 (2025~\cite{b200-ga}), via distinct mechanisms.
For Qwen 3 32B, H100's 80 GB VRAM cannot fit BF16 weights plus the KV cache beyond batch size $32$, lowering the ceiling and giving a narrow feasible power range; B200's 192 GB lifts the memory cap, raising the ceiling and delivering a wider range.
For Qwen 3 8B, both GPUs admit the same feasible batch range, but B200 has higher compute throughput, so a larger fraction of its capacity sits idle at small batches than H100, widening the gap between small-batch and peak power and therefore the feasible power range.

\emph{Takeaway.}
AI model choices (e.g., size, architecture, precision) and deployment choices (e.g., parallelism, hardware) shape the feasible power range by lifting or lowering its endpoints.
Model-side, MoE and FP8 can lift the floor while a tight latency target caps the ceiling; deployment-side, larger parallelism can lower the floor, and a newer GPU can either lower the floor (more compute headroom) or raise the ceiling (relieved memory bottleneck).
Configurations with wider feasible power range (e.g., Qwen 3 8B on B200) give controllers more room to mitigate grid stress and reduce integral voltage violation than configurations with narrower range (e.g., Qwen 3 32B on H100).

\section{Related Work}\label{sec:related}

\paragraph{Model benchmarks.}
A growing line of work measures energy and performance of generative AI across hardware and serving configurations~\cite{mlenergy-benchmark-neurips25,mlenergy-benchmark-v3-arxiv26,mlperf-power-hpca25,google-ai-energy-arxiv25,ai-energy-score}.
OpenG2G's default datacenter backend uses the ML.ENERGY Benchmark~\cite{mlenergy-benchmark-neurips25,mlenergy-benchmark-v3-arxiv26} because it covers numerous relevant models with openly available, production-grade measurements that jointly report inference time, throughput, and power.
These benchmarks help characterize the impact of important knobs across the stack on AI model service metrics, but grid and datacenter control is outside their scope.

\paragraph{Datacenter and grid control.}
Datacenter-focused simulators~\cite{sustaindc-neuripsdb24,pydcm-buildsys23} model the internals of the datacenter, yet are agnostic to the grid.
On the other hand, conventional grid research~\cite{opendssdirect,grid2op,powergym-arxiv21,rl2grid-arxiv25,opfgym-energyai24} are either agnostic to datacenters or treat them as simple loads. More recent work has explored the use of datacenters' flexibility for optimization and control of grid-side metrics~\cite{fu2020assessments,agentconcur-tpwrs25,xie2025enhancing,chen2025voltage,colangelo2025ai,g2g-powerup26}, signifying a growing interest in the coordination problem.
However, these works are fragmented in the granularity of datacenter modeling, and vary significantly in terms of workload realism and coordination strategies, making it difficult to compare them on a common basis.
OpenG2G is, to the best of our knowledge, the first open-source framework for simulating the runtime interaction between an AI datacenter and its host grid.
With its modular design and support for realistic workloads, OpenG2G enables the comparison of various solutions to the energy challenge of AI on a common basis, thus bridging datacenter and grid communities and offering a platform for research at their intersection.

\section{Conclusion}\label{sec:conclusion}

AI datacenters are now large and dynamic enough to impact the grid, making the coordination between the two a control problem.
OpenG2G makes this problem approachable behind a single Python interface: a datacenter backend driven by measurements of realistic AI services, a grid backend wrapping a traditional grid simulator, and a controller interface behind which classical feedback and learning-based policies plug in on equal footing.
Using OpenG2G, we show that AI model choices (e.g., size, architecture, precision) and deployment choices (e.g., parallelism, hardware) each project onto a measurable feasible power range and a corresponding headroom for voltage regulation, so the space of viable datacenter configurations is inseparable from the grid it lands on.
In the era of gigawatt-scale AI infrastructure, we hope to foster a rich community of researchers and practitioners across machine learning, systems, and power systems around OpenG2G, together turning AI datacenter--grid runtime coordination from a pressing concern into a tractable, well-understood research area.

\bibliographystyle{plain}
\bibliography{ref}

\clearpage
\appendix
\section{Extending OpenG2G}\label{sec:appendix-extensibility}

OpenG2G extension points are ordinary Python classes or data objects; the four listings below illustrate the most common extensions.

\paragraph{Adding a measured inference workload (Listing~\ref{lst:appendix-workload-spec}).}
OpenG2G's datacenter component plays back per-batch power, throughput, and inter-token-latency distributions measured by the ML.ENERGY Benchmark, so adding a new inference workload reduces to declaring a model specification: a Python data object that identifies the measurement slice (model id, GPU, task, precision, parallelism, GPUs per replica) and the serving-level constraints (latency target, feasible batch sizes).
The spec doubles as a content-addressed cache key, so two scripts referring to the same logical configuration share one on-disk cache.
Workloads outside ML.ENERGY plug in without modifying OpenG2G: a user can either populate the same per-spec on-disk schema (a \texttt{trace.csv} of $(\text{batch size}, \text{relative time}, \text{power})$ rows and an \texttt{itl\_fit.json} of fitted ITL parameters), or assemble the runtime data objects (\texttt{InferenceTraceStore} and \texttt{ITLFitStore}) directly in Python from custom inputs.
For workloads with no measurements at all, a custom \texttt{DatacenterBackend} subclass can replace offline replay entirely and emit power and service state directly (the same pattern as the custom grid backend in Listing~\ref{lst:appendix-grid}).
A deployment object then pairs a spec with runtime state such as the initial batch size, letting the same spec be instantiated multiple times within a single scenario.

\paragraph{Implementing a datacenter-side controller (Listing~\ref{lst:appendix-controller}).}
A controller subclasses the abstract \texttt{Controller} base and implements a single \texttt{step} method that reads datacenter and grid state and returns a list of typed commands.
The example \texttt{DeadbandController} shows the minimal shape: it reads the bus-voltage vector from the grid handle, halves the current batch size on undervoltage, doubles it on overvoltage, and otherwise emits no command (roughly thirty lines including imports).
The controller's native cadence is set by the \texttt{dt\_s} property (here, every 1 s); the simulation loop advances the controller at this rate independently of the datacenter and grid simulation cadences.
The returned \texttt{SetBatchSize} command is typed and routed to the target datacenter, so a single controller can act on multiple datacenters or mix datacenter and grid commands in one \texttt{step} call.

\paragraph{Implementing a grid-side controller (Listing~\ref{lst:appendix-adaptive-tap}).}
Grid-side controllers use the same interface, differing only in the commands they emit.
\texttt{AdaptiveTapController} observes feeder voltages, picks a regulator and a direction based on the worst margin, and adjusts the tap by one physical step after a cooldown interval; the cooldown reflects the mechanical reality that on-load tap changers cannot be operated too frequently without wear.
Its native cadence is 5 s, slower than a typical batch size controller, illustrating the framework's multi-rate scheduling.
The emitted \texttt{SetTaps} command is typed; the simulation loop routes it to the grid backend.

\paragraph{Custom grid backend (Listing~\ref{lst:appendix-grid}).}
The default grid backend wraps OpenDSS, but any subclass of \texttt{GridBackend} that consumes datacenter power samples and returns OpenG2G grid state can plug in.
At each tick the backend receives a dictionary of per-datacenter three-phase power samples accumulated over the past simulation step, hands them to a user-provided solver, and returns a \texttt{GridState} containing per-bus voltages and tap positions.
This is the seam through which a real-time co-simulation harness or a site-specific feeder model can be substituted without changing controllers, datacenters, or scenario configurations.

\paragraph{Type-safe composition.}
OpenG2G uses Python's generics to encode component compatibility at the type level: \texttt{Controller} is parameterized as \texttt{Controller[DCBackendT, GridBackendT]}, and a concrete controller specializes those parameters to declare which datacenter and grid types it can act on.
For example, \texttt{DeadbandController} above specializes to \texttt{Controller[LLMBatchSizeControlledDatacenter, OpenDSSGrid]}, which lets a static type checker verify that \texttt{self.dc.state.batch\_size\_by\_model} resolves to an actual field. This field lives on \texttt{LLMDatacenterState}, not on the universal \texttt{DatacenterState} base.
Datacenter and grid backends are themselves generic over their state types (\texttt{DatacenterBackend[DCStateT]}, \texttt{GridBackend[GridStateT]}), so a backend that emits a richer state type automatically exposes the richer fields to controllers bound to the same parameterization, with no runtime casts and no \texttt{hasattr} checks.
The same parameterization is enforced at scenario assembly time: instantiating a controller against a datacenter or grid that does not satisfy its bounds raises a typed error rather than failing midway through a long simulation run.

\begin{listing}[t]
\begin{tcolorbox}[
  enhanced,
  colback=codebg,
  colframe=gray!40,
  boxrule=0.4pt,
  arc=1.5mm,
  left=2mm, right=2mm, top=0mm, bottom=1mm,
  title={Measured inference workload},
  coltitle=black!90,
  fonttitle=\footnotesize\bfseries,
  colbacktitle=gray!20,
  bottomtitle=0.4pt,
  titlerule=0pt,
]
\begin{minted}{python}
from openg2g.datacenter.config import InferenceModelSpec, ModelDeployment

qwen = InferenceModelSpec(
    model_label="Qwen3-32B-B200",
    model_id="Qwen/Qwen3-32B",
    gpu_model="B200",
    task="lm-arena-chat",
    precision="bfloat16",
    gpus_per_replica=1,
    tensor_parallel=1,
    itl_deadline_s=0.050,
    batch_sizes=(8, 16, 32, 64, 128, 256, 512),
    feasible_batch_sizes=(8, 16, 32, 64, 128),
)

deployment = ModelDeployment(spec=qwen, initial_batch_size=128)
\end{minted}
\end{tcolorbox}
\caption{Adding a measured inference workload. The model spec identifies the ML.ENERGY measurement slice and the serving-level constraints; deployment-specific choices such as the initial batch size are kept separate.}
\label{lst:appendix-workload-spec}
\end{listing}

\begin{listing}[H]
\begin{tcolorbox}[
  enhanced,
  colback=codebg,
  colframe=gray!40,
  boxrule=0.4pt,
  arc=1.5mm,
  left=2mm, right=2mm, top=0mm, bottom=1mm,
  title={Datacenter-side controller},
  coltitle=black!90,
  fonttitle=\footnotesize\bfseries,
  colbacktitle=gray!20,
  bottomtitle=0.4pt,
  titlerule=0pt,
]
\begin{minted}{python}
from fractions import Fraction

from openg2g.clock import SimulationClock
from openg2g.controller.base import Controller
from openg2g.datacenter.base import LLMBatchSizeControlledDatacenter
from openg2g.datacenter.command import DatacenterCommand, SetBatchSize
from openg2g.events import EventEmitter
from openg2g.grid.command import GridCommand
from openg2g.grid.opendss import OpenDSSGrid


class DeadbandController(Controller[LLMBatchSizeControlledDatacenter, OpenDSSGrid]):
    def __init__(
        self,
        dc: LLMBatchSizeControlledDatacenter,
        grid: OpenDSSGrid,
        model_label: str,
    ) -> None:
        self.dc = dc
        self.grid = grid
        self.model_label = model_label

    @property
    def dt_s(self) -> Fraction:
        return Fraction(1, 1)

    def reset(self) -> None:
        pass

    def step(
        self,
        clock: SimulationClock,
        events: EventEmitter,
    ) -> list[DatacenterCommand | GridCommand]:
        v = self.grid.voltages_vector()
        current = self.dc.state.batch_size_by_model[self.model_label]
        if v.min() < 0.95:
            target = max(8, current // 2)
        elif v.max() > 1.05:
            target = min(512, current * 2)
        else:
            return []
        return [SetBatchSize({self.model_label: target}, target=self.dc)]
\end{minted}
\end{tcolorbox}
\caption{A minimal datacenter-side controller. The simulation loop calls \texttt{step} at the controller's native cadence and routes the returned \texttt{SetBatchSize} command to the target datacenter.}
\label{lst:appendix-controller}
\end{listing}

\begin{listing}[H]
\begin{tcolorbox}[
  enhanced,
  colback=codebg,
  colframe=gray!40,
  boxrule=0.4pt,
  arc=1.5mm,
  left=2mm, right=2mm, top=0mm, bottom=1mm,
  title={Adaptive tap-position controller},
  coltitle=black!90,
  fonttitle=\footnotesize\bfseries,
  colbacktitle=gray!20,
  bottomtitle=0.4pt,
  titlerule=0pt,
]
\begin{minted}{python}
from fractions import Fraction

import numpy as np

from openg2g.clock import SimulationClock
from openg2g.controller.base import Controller
from openg2g.datacenter.base import DatacenterBackend
from openg2g.datacenter.command import DatacenterCommand
from openg2g.events import EventEmitter
from openg2g.grid.command import GridCommand, SetTaps
from openg2g.grid.config import TapPosition
from openg2g.grid.opendss import OpenDSSGrid


class AdaptiveTapController(Controller[DatacenterBackend, OpenDSSGrid]):
    def __init__(
        self,
        grid: OpenDSSGrid,
        *,
        regulator: str,
        tap_step: float = 0.00625,
        tap_min: float = 0.90,
        tap_max: float = 1.10,
        deadband: float = 0.002,
        cooldown_s: float = 60.0,
    ) -> None:
        self.grid = grid
        self.regulator = regulator
        self.tap_step = tap_step
        self.tap_min = tap_min
        self.tap_max = tap_max
        self.deadband = deadband
        self.cooldown_s = cooldown_s
        self.next_allowed_s = 0.0

    @property
    def dt_s(self) -> Fraction:
        return Fraction(5, 1)

    def reset(self) -> None:
        self.next_allowed_s = 0.0

    def step(
        self,
        clock: SimulationClock,
        events: EventEmitter,
    ) -> list[DatacenterCommand | GridCommand]:
        if clock.time_s < self.next_allowed_s:
            return []

        v = self.grid.voltages_vector()
        low_margin = 0.95 - np.nanmin(v)
        high_margin = np.nanmax(v) - 1.05
        if low_margin <= self.deadband and high_margin <= self.deadband:
            return []

        tap_state = self.grid.state.tap_positions
        taps = tap_state.regulators if tap_state is not None else {}
        current = taps.get(self.regulator.lower(), 1.0)
        direction = 1 if low_margin > high_margin else -1
        target = current + direction * self.tap_step
        target = min(self.tap_max, max(self.tap_min, target))
        if target == current:
            return []

        self.next_allowed_s = clock.time_s + self.cooldown_s
        return [SetTaps(tap_position=TapPosition(regulators={self.regulator: target}))]
\end{minted}
\end{tcolorbox}
\caption{An adaptive tap-position controller. It observes feeder voltages, changes one regulator by at most one physical tap step after a cooldown interval, and emits a \texttt{SetTaps} grid command.}
\label{lst:appendix-adaptive-tap}
\end{listing}

\begin{listing}[H]
\begin{tcolorbox}[
  enhanced,
  colback=codebg,
  colframe=gray!40,
  boxrule=0.4pt,
  arc=1.5mm,
  left=2mm, right=2mm, top=0mm, bottom=1mm,
  title={Custom grid backend},
  coltitle=black!90,
  fonttitle=\footnotesize\bfseries,
  colbacktitle=gray!20,
  bottomtitle=0.4pt,
  titlerule=0pt,
]
\begin{minted}{python}
from fractions import Fraction
from typing import Any

import numpy as np

from openg2g.clock import SimulationClock
from openg2g.common import ThreePhase
from openg2g.datacenter.base import DatacenterBackend
from openg2g.events import EventEmitter
from openg2g.grid.base import (
    BusVoltages,
    GridBackend,
    GridState,
    PhaseVoltages,
)
from openg2g.grid.command import GridCommand

PowerSamples = dict[DatacenterBackend, list[ThreePhase]]


class CoSimulationGrid(GridBackend[GridState]):
    def __init__(self, solver: Any, dc_bus: str = "671") -> None:
        super().__init__()
        self.solver = solver
        self.dc_bus = dc_bus

    @property
    def dt_s(self) -> Fraction:
        return Fraction(1, 10)

    def reset(self) -> None:
        self.solver.reset()

    def step(
        self,
        clock: SimulationClock,
        power_samples_w: PowerSamples,
        events: EventEmitter,
    ) -> GridState:
        samples = next(iter(power_samples_w.values()), [])
        p_w = np.mean([p.a + p.b + p.c for p in samples])
        raw = self.solver.solve_load(self.dc_bus, p_w)
        voltages = BusVoltages({
            bus: PhaseVoltages(v[0], v[1], v[2])
            for bus, v in raw.voltages.items()
        })
        return GridState(clock.time_s, voltages, raw.tap_positions)

    def apply_control(self, command: GridCommand, events: EventEmitter) -> None:
        self.solver.apply(command)
\end{minted}
\end{tcolorbox}
\caption{Sketch of a custom grid adapter showing the load$\to$voltage seam: a backend can wrap OpenDSS, a co-simulator, or a site-specific solver as long as it consumes datacenter power samples and returns OpenG2G grid state. A runnable subclass additionally implements \texttt{voltages\_vector}, \texttt{v\_index}, and \texttt{estimate\_sensitivity}, used by controllers that probe voltage sensitivity such as OFO.}
\label{lst:appendix-grid}
\end{listing}

\section{Mathematical Foundation of Controllers}\label{sec:appendix-controller-math}
\subsection{Droop control}
The droop controller is a discrete-time proportional law that adjusts each model's batch size in response to the worst observed voltage violation. At each control step $t$, let ${\mathbf v}(t) \in \mathbb{R}^{3M}$ denote the bus-voltage vector and $[ \underline v, \overline v]$ the allowable band (we use $[0.95, 1.05]$ pu), where $M$ is the total number of buses in the system. The dimension of ${\mathbf v}(t)$ is $3M$ rather than $M$ because each bus has three phases $\phi\in\{A,B,C\}$, with possibly different voltage measurement in each phase. Define the unsigned violation magnitudes
\[
  \varepsilon_-(t) = \max\bigl(0, \underline v - \min_{j,\phi} v_{j,\phi}(t)\bigr),
  \qquad
  \varepsilon_+(t) = \max\bigl(0, \max_{j,\phi} v_{j,\phi}(t) - \overline v\bigr),
\]
and a signed scalar pressure $p(t) = \varepsilon_-(t) - \varepsilon_+(t)$ that is zeroed inside a deadband $|p(t)| \le \delta$ to suppress chattering. Here, $v_{j,\phi}$ is the voltage measurement at bus $j$ and phase $\phi$. For each LLM $i$ deployed at the datacenter, the controller maintains a continuous log-domain state $x_i(t) = \log_2 b_i(t)$ and updates it by
\[
x_i(t+1)
=
\Pi_{[\underline{x}_i,\overline{x}_i]}
\Bigl(x_i(t) - K_p p(t) \Bigl), \qquad
b_i(t+1)
= \arg\min_{b\in\mathcal{B}_i}
\left| b - 2^{x_i(t+1)} \right|.
\]
where $\mathcal{B}_i$ is the model's feasible batch size set and $K_p$ is a single proportional gain (step-size in the implementation), and $\Pi_{[\underline{x}_i,\overline{x}_i]}(\cdot)$ denotes projection onto the constraints $\underline x_i \le x_i(t) \le \overline x_i$. A positive pressure (undervoltage) shrinks $x_i$, reducing power draw; a negative pressure (overvoltage) grows it. A latency guard suppresses any increase whenever the observed inter-token latency $\mathrm{ITL}_i(t)$ exceeds the model's latency target. The droop controller is purely local to the datacenter, because it requires neither a sensitivity matrix nor any model fits. It serves as the simplest baseline against which OFO and PPO are compared. In multi-datacenter feeders, each site runs an independent instance with the same $K_p$; all instances observe the same global bus-voltage vector and therefore react in lockstep to the worst violation in the network. In this paper, we use $K_p = 50$ in all the experiments.

\subsection{Online feedback optimization (OFO) control}
The Online Feedback Optimization (OFO) controller is a primal--dual method that solves a constrained Lagrangian on the running grid state and converges to a KKT point of an equivalent constrained optimization without ever requiring a forward model of the network~\cite{g2g-powerup26}. We treat the per-step batch problem as
\[
\begin{aligned}
\max_{\{x_i\}}\; & \;\alpha_T\sum_{i} T_i(x_i) \;-\; \beta_S \sum_{i}\bigl(x_i - x_i(t{-}1)\bigr)^2 \\
\text{s.t.}\; & \;\underline{v}\;\le\; v_{j,\phi}\bigl(\mathbf{p}(\{x_i\})\bigr)\;\le\;\overline{v} && \forall j,\phi,\\
& \;L_i(x_i)\;\le\;\mathrm{ITL}_i^{\,\star} && \forall i,
\end{aligned}
\]
where $T_i(\cdot), L_i(\cdot), P_i(\cdot)$ are real-time measurement of token throughput, inter-token latency, and datacenter power with the log-batch state $x_i$; $\mathrm{ITL}_i^{\,\star}$ is model $i$'s latency target; $\alpha_T$ trades throughput against constraint slack; and $\beta_S$ regularizes the switching cost. Voltage dependence on batch state is mediated by the bus-injection sensitivity $\mathbf{H} = \partial \mathbf{v}/\partial \mathbf{p} \in \mathbb{R}^{3M\times 3}$ at the datacenter bus, estimated online via finite-difference probes every $\tau_H$ control steps.

Let $\underline{\boldsymbol{\lambda}}, \overline{\boldsymbol{\lambda}}\in\mathbb{R}^{3M}_+$ be the dual multipliers for the lower and upper voltage bounds and $\boldsymbol{\eta}(t) = \overline{\boldsymbol{\lambda}}(t) - \underline{\boldsymbol{\lambda}}(t)$ their signed difference, and let $\mu_i\ge 0$ be the latency dual for model $i$. At each control step the controller performs a single first-order primal--dual update:
\[
\begin{aligned}
\underline{\boldsymbol{\lambda}}(t{+}1) &= \bigl[\,\underline{\boldsymbol{\lambda}}(t) + \rho_v\bigl(\underline{v}\,\mathbf{1} - \mathbf{v}(t)\bigr)\bigr]_+, \\
\overline{\boldsymbol{\lambda}}(t{+}1) &= \bigl[\,\overline{\boldsymbol{\lambda}}(t) + \rho_v\bigl(\mathbf{v}(t) - \overline{v}\,\mathbf{1}\bigr)\bigr]_+, \\
\mu_i(t{+}1) &= \bigl[\,\mu_i(t) + \rho_l\bigl(\mathrm{ITL}_i(t) - \mathrm{ITL}_i^{\,\star}\bigr)\bigr]_+, \\
x_i(t{+}1) &= \Pi_{[\underline{x}_i,\overline{x}_i]}\!\Bigl(x_i(t) - \rho_x\,\nabla_{x_i}\mathcal{L}(t)\Bigr), \\
b_i(t{+}1) &= \arg\min_{b\in\mathcal{B}_i}\bigl|\,b - 2^{x_i(t{+}1)}\bigr|,
\end{aligned}
\]
where $[\,\cdot\,]_+ = \max(\,\cdot\,;\,0)$ enforces dual non-negativity, and the primal Lagrangian gradient is
\[
\nabla_{x_i}\mathcal{L}(t)
\;=\;
-\,\alpha_T\,\frac{\partial T_i}{\partial x_i}
\;+\; k_v\,\bigl(\boldsymbol{\eta}(t)^{\!\top} \mathbf{H}\,\mathbf{e}_i\bigr)\,\frac{\partial P_i}{\partial x_i}
\;+\; \mu_i(t)\,\frac{\partial L_i}{\partial x_i}
\;+\; 2\beta_S\bigl(x_i(t) - x_i(t{-}1)\bigr).
\]
Here $\mathbf{e}_i\in\Delta^2$ is the per-model phase-allocation share at the datacenter bus and $k_v$ is a fixed scaling coefficient that converts the dual--sensitivity inner product into a power-domain gradient; the gradient of $T_i$, $P_i$ and $L_i$ with respect to $x_i$ can be estimated using logistic fits of throughput, power, and latency against $x_i$; we refer the reader to~\cite{g2g-powerup26} for the full derivation of the gradient. Both objective weights $(\alpha_T, \beta_S)$ are tunable: we set them uniformly to $\alpha_T = 10^{-4}$ and $\beta_S = 1.0$ as a default across all three feeders so that the throughput term and switching regularization remain comparable in magnitude to the constraint-satisfaction terms once duals settle, and we study the impact of $\alpha_T$ specifically in the throughput--voltage trade-off sensitivity analysis (Appendix~\ref{sec:appendix-controller-perf}). The remaining hyperparameters, including the primal step size $\rho_x$, the dual step sizes $(\rho_v, \rho_l)$, the sensitivity-scaling coefficient $k_v$, the sensitivity-update period $\tau_H$, and the finite-difference probe magnitude $\delta_p$, are listed per feeder in Table~\ref{tab:ofo_hyperparams}. The probe magnitude $\delta_p$ is chosen system-by-system to balance two competing requirements: it must be small enough that the load-flow response to a $\delta_p$-kW perturbation at the datacenter bus stays in the locally-linear regime so that the finite-difference estimate of $\mathbf{H} = \partial \mathbf{v}/\partial \mathbf{p}$ is unbiased, and large enough that the resulting voltage signal at remote bus-phases exceeds simulator and floating-point noise. Empirically the right scale for $\delta_p$ tracks the typical control-induced power swing at the datacenter bus, which scales with the site's GPU capacity and the natural amplitude of PV/TVL disturbances on the feeder; we use $\delta_p \in \{100, 50, 10\}$~kW for IEEE-13, IEEE-34, and IEEE-123 respectively, with the smallest value chosen on the largest feeder where stronger inter-bus coupling makes a small probe sufficient to recover $\mathbf{H}$.
\begin{table}[t]
  \centering
  \caption{OFO controller hyperparameters per feeder. The objective weights $(\alpha_T,\beta_S)$ and the sensitivity-update period $\tau_H$ are held constant across feeders; only the dual step sizes and the finite-difference probe magnitude $\delta_p$ used to estimate $\mathbf{H}$ vary per system.}
  \label{tab:ofo_hyperparams}
  \small
  \setlength{\tabcolsep}{6pt}
  \begin{tabular}{l c c c c c c c c}
    \toprule
    System & $\alpha_T$ & $\beta_S$ & $k_v$ & $\rho_x$ & $\rho_v$ & $\rho_l$ & $\tau_H$ (s) & $\delta_p$ (kW) \\
    \midrule
    IEEE-13  & $10^{-4}$ & $1.0$ & $10^{6}$ & $0.05$ & $1.0$ & $1.0$ & $300$ & $100$ \\
    IEEE-34  & $10^{-4}$ & $1.0$ & $10^{6}$ & $0.05$ & $1.0$ & $1.0$ & $300$ & $50$  \\
    IEEE-123 & $10^{-4}$ & $1.0$ & $10^{6}$ & $0.05$ & $0.3$ & $1.0$ & $300$ & $10$  \\
    \bottomrule
  \end{tabular}
\end{table}

OFO requires explicit logistic fits of $P_i, T_i, L_i$ -- obtained from offline batch size sweeps on the deployed inference servers -- and online estimation of $\mathbf{H}$. In return it provides a closed-loop primal--dual optimization with a KKT-stationarity guarantee under standard conditions; we use it as the strongest model-based reference against which the model-free PPO controller is compared.

\subsection{Proximal Policy Optimization (PPO) control}
The PPO controller learns a closed-loop stochastic policy $\pi_\theta(\boldsymbol{\Delta}| \mathbf{o})$, parameterized by neural-network weights $\theta$, that maps a system observation $\mathbf{o}(t)$ to a vector of per-model batch size adjustments $\boldsymbol{\Delta}(t)$. PPO~\cite{ppo-arxiv17} is a model-free, on-policy actor--critic method: it alternates between (i) collecting on-policy trajectories of $(\mathbf{o}_t,\boldsymbol{\Delta}_t,r_t)$ tuples under the current data-collecting policy $\pi_{\theta_{\mathrm{old}}}$, and (ii) updating the parameters $\theta$ by stochastic gradient ascent on the clipped surrogate objective
\[
L^{\mathrm{CLIP}}(\theta)
= \mathbb{E}_t\!\Bigl[\min\bigl(\rho_t(\theta)\,\hat A_t,\;
\Pi_{[1{-}\epsilon,1{+}\epsilon]}(\rho_t(\theta))\,\hat A_t\bigr)\Bigr],
\qquad
\rho_t(\theta) = \frac{\pi_\theta(\boldsymbol{a}_t\mid \mathbf{o}_t)}{\pi_{\theta_{\mathrm{old}}}(\boldsymbol{a}_t\mid
\mathbf{o}_t)}.
\]
where $\boldsymbol{\Delta}_t$ is the action sampled from $\pi_{\theta_{\mathrm{old}}}$ at step $t$, $\rho_t(\theta)$ is the likelihood ratio of the candidate policy $\pi_\theta$ to the data-collecting policy $\pi_{\theta_{\mathrm{old}}}$ on that action, $\epsilon\in(0,1)$ is a clipping hyperparameter (we use $\epsilon=0.2$), and $\hat A_t$ is the estimated advantage of taking action $\boldsymbol{\Delta}_t$ in observation $\mathbf{o}_t$ -- computed by generalized advantage estimation (GAE)~\cite{schulman2015high} from the trajectory rewards $\{r_\tau\}_{\tau\ge t}$ and a learned value baseline $V_\phi(\mathbf{o})$ produced by a separate critic network with parameters $\phi$. The $\mathrm{clip}$ term restricts each policy update to a trust region around $\pi_{\theta_{\mathrm{old}}}$, which is the source of PPO's training stability. Compared with off-policy actor--critic methods such as DDPG and SAC that learn from a replay buffer, PPO trades sample efficiency for training stability -- well suited to our non-stationary multi-scenario training distribution. Compared with value-based methods such as DQN, PPO directly parameterizes a stochastic policy and accommodates joint multi-dimensional action spaces (one categorical distribution per LLM here) without the combinatorial $|\mathcal{A}|$-way value head that DQN-family methods would require.

\paragraph*{Observation space.}
The observation $\mathbf{o}(t)$ has two parts: a voltage block whose form depends on the \texttt{--obs-mode} flag, and a per-LLM block that is identical across modes. The per-LLM block contributes $5$ scalars per deployed LLM: the normalized log-batch state $(\log_2 b_i(t) - \log_2 b_i^{\min})/(\log_2 b_i^{\max} - \log_2 b_i^{\min})$, the observed inter-token latency normalized by the latency target $\mathrm{ITL}_i(t)/\mathrm{ITL}_i^{\,\star}$ (clipped to $[0, 3]$), the active replica fraction $r_i(t)/r_i^{\max}$, the total datacenter power in MW (shared across LLM entries), and the previous-step log-batch delta $(\log_2 b_i(t) - \log_2 b_i(t{-}1))$ rescaled to $[-1, 1]$. Table~\ref{tab:obs_modes} lists the three voltage encodings used in the paper.

\begin{table}[t]
  \centering
  \small
  \caption{Observation modes (\texttt{--obs-mode} flag). Every mode shares the same $5N$ per-LLM features; modes differ in how voltage information is summarized. $M$ is the number of bus-phases ($M = 3 M_\mathrm{bus}$ in our three-phase feeders), $Z$ the number of zones, $N$ the number of deployed LLMs. The 3-scalar global summary block is $(\max\text{ under-violation},\,\max\text{ over-violation},\,\text{frac.\ in violation})$; the per-zone variant replicates these three statistics per zone.}
  \label{tab:obs_modes}
  \begin{tabular}{lcccc}
    \toprule
    \texttt{--obs-mode} & Voltage block & Summary block & Total dim & Used by \\
    \midrule
    \texttt{full-voltage}        & $\mathbf{v}(t)\in\mathbb{R}^{M}$              & $3$ global  & $M + 3 + 5N$                 & IEEE-13 \\
    \texttt{per-zone-summary}    & ---                                            & $3Z$        & $3Z + 5N$                    & IEEE-123 \\
    \texttt{system-summary-only} & ---                                            & $3$ global  & $3 + 5N$                     & IEEE-34 \\
    \bottomrule
  \end{tabular}
\end{table}

The three modes correspond to progressively coarser views of the voltage state. \texttt{full-voltage} exposes every bus-phase voltage and is suitable for small feeders. \texttt{per-zone-summary} keeps only the per-zone violation statistics (requires zones to be defined). And \texttt{system-summary-only} retains only three scalars, decoupling the policy's input dimension from the size of the feeder. We use the mode listed in the right column for each system in our main experiments.

\paragraph*{Action space.} At each control step $t$ the policy receives an observation $\mathbf{o}(t)$ that concatenates a voltage feature vector -- either the full bus-voltage measurement $\mathbf{v}(t)\in\mathbb{R}^{3M}$, a per-zone summary, or a system-wide summary $(\min,\, \max,\, \mathrm{mean})$ -- together with the current per-model log-batch state $\{x_i(t)\}_{i=1}^N$. The action is a discrete vector $\boldsymbol{\Delta}(t)\in\{-1,\,0,\,+1\}^N$ over the $N$ deployed LLMs (``decrement / hold / increment''), and the policy is parameterized as $N$ independent categorical distributions, one per LLM. Let $\mathrm{idx}_i(t)\in\{0,\ldots,|\mathcal{B}_i|-1\}$ index the model's ordered feasible batch size set $\mathcal{B}_i$; the batch update is
\[
\begin{aligned}
\mathrm{idx}_i(t+1)
&= \Pi_{[0,|\mathcal{B}_i|-1]}\!\bigl(\mathrm{idx}_i(t) + \Delta_i(t) \bigr), \\
b_i(t+1)
&= \mathcal{B}_i\bigl[\mathrm{idx}_i(t+1)\bigr],
\qquad
x_i(t+1) = \log_2 b_i(t+1).
\end{aligned}
\]
This delta parameterization shrinks the per-LLM action from $|\mathcal{B}_i|$ absolute choices to $3$ relative ones, reducing the joint action space from $\prod_i |\mathcal{B}_i|$ (e.g.\ $7^N$ when each $\mathcal{B}_i$ has 7 levels) to $3^N$ -- a key enabler of tractable PPO exploration on multi-LLM datacenters. 

\paragraph*{Reward function.}
The agent receives a scalar reward
\[
r(t) = -\,w_V\,\mathcal{P}_V(t) \;+\; w_T\,\mathcal{T}(t)
       \;-\; w_L\,\mathcal{P}_L(t) \;-\; w_S\,\mathcal{P}_S(t).
\]
The voltage-violation penalty is the squared bus-phase exceedance summed across the grid,
\[
\mathcal{P}_V(t) = \sum_{j,\phi}\Bigl[\,\max\!\bigl(0;\, \underline{v} - v_{j,\phi}(t)\bigr)^{2}
                                     + \max\!\bigl(0;\, v_{j,\phi}(t) - \overline{v}\bigr)^{2}\,\Bigr].
\]
The throughput bonus is the per-model token throughput summed across the deployment and normalized by each model's peak so it lies in $[0, 1]$,
\[
\mathcal{T}(t) = \sum_{i=1}^{N}\,\frac{T_i\!\bigl(b_i(t)\bigr)}{T_i\!\bigl(b_i^{\max}\bigr)},
\qquad b_i^{\max} = \max\,\mathcal{B}_i.
\]
We compute $T_i(b_i(t))$ via the same offline logistic fit used by OFO, but only because our simulator does not expose a measurable throughput counter; in a real deployment $T_i(b_i(t))$ would be read directly from the inference server's telemetry (tokens/sec emitted by, e.g., vLLM or TGI), so the throughput fit is a simulator stand-in rather than a deployment-time dependency of the PPO controller.

The latency penalty is the relative excess of the observed inter-token latency $\mathrm{ITL}_i(t)$ over each model's latency target $\mathrm{ITL}_i^{\,\star}$ (only positive excesses contribute),
\[
\mathcal{P}_L(t) = \sum_{i=1}^{N}\,\max\!\Bigl(0;\;\frac{\mathrm{ITL}_i(t) - \mathrm{ITL}_i^{\,\star}}{\mathrm{ITL}_i^{\,\star}}\Bigr).
\]
The switching penalty is the log-domain magnitude of any batch size change since the previous step,
\[
\mathcal{P}_S(t) = \bigl|\,\log_2 b_i(t) - \log_2 b_i(t{-}1)\,\bigr|.
\]
The four weights $(w_V, w_T, w_L, w_S)$ are tuned per system. The weights used by the best-performed PPO models are listed in Table~\ref{tab:ppo_hyperparams}.

\paragraph*{Training and deployment.} We train a single shared policy per feeder with stable-baselines3's PPO implementation, using three-layer MLP actor and critic networks of width $128$, observation normalization (running-statistics whitening), $n_{\mathrm{envs}}=8$ parallel rollout workers, $n_{\mathrm{steps}}=3600$, entropy coefficient $0.01$, GAE with $\gamma=0.99$, $\lambda=0.95$, and clip range $\epsilon=0.2$, for $2$--$5\times10^6$ environment steps drawn from a randomized scenario library. Unlike OFO, in which the throughput, latency, and power fits $T_i, L_i, P_i$ together with the online voltage-sensitivity estimate $\mathbf{H}$ are inputs to the closed-loop Lagrangian gradient at every deployment-time control step, the trained PPO policy depends on no explicit model of the coupled LLM/grid dynamics: the action-conditional structure has been absorbed into the network weights during training. Throughput, inter-token latency, and bus voltage enter PPO only as measurements: they appear in the training reward through whatever sensor or simulator is providing the rollouts. At deployment, a single forward pass through the policy network on the observation $\mathbf{o}(t)$ produces the batch size adjustments for every deployed LLM, with no online learning, no online probing, and no model-fit dependency.

\begin{table}[t]
      \centering
      \caption{PPO champion training hyperparameters per feeder. Per-system reward weights are over voltage-violation integral $w_V$, throughput $w_T$, latency $w_L$, and switching cost $w_S$ as defined in the reward equation; \emph{Best ckpt} reports the checkpoint that minimized mean integral on the 50-scenario test set; observation dimensions follow Table~\ref{tab:obs_modes}.}
      \label{tab:ppo_hyperparams}
      \small
      \setlength{\tabcolsep}{4pt}
      \begin{tabular}{l c c c c r r r r c c c}
        \toprule
        System & $N_\mathrm{LLM}$ & $N_\mathrm{DC}$ & Lib size & lr & $w_V$ & $w_T$ & $w_L$ & $w_S$ & Steps & Seed &
  Best ckpt \\
        \midrule
        IEEE-13  & 5 & 1 & 235 & $1{\cdot}10^{-4}$ & 5000 & $5{\cdot}10^{-2}$ & $1{\cdot}10^{-2}$ & 0.5 & 2.0M & 1 &
  1.44M \\
        IEEE-34  & 5 & 2 & 168 & $5{\cdot}10^{-5}$ & 5000 & $2{\cdot}10^{-4}$ & $1{\cdot}10^{-3}$ & 1.0 & 5.0M & 5 &
  3.74M \\
        IEEE-123 & 4 & 4 & 231 & $1{\cdot}10^{-4}$ & 5000 & $1{\cdot}10^{-2}$ & $1{\cdot}10^{-1}$ & 1.0 & 5.0M & 5 &
  3.74M \\
        \bottomrule
      \end{tabular}
    \end{table}

\section{IEEE Test Feeders}\label{sec:appendix-grid}
\begin{figure*}[htbp]
    \centering

    \includegraphics[width=\textwidth]{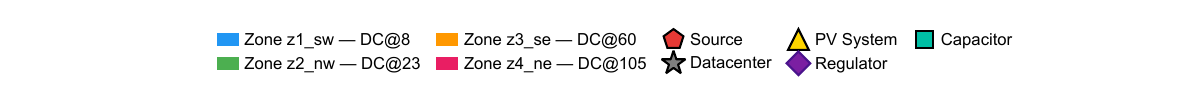}
    \vspace{0.3em}
    \begin{minipage}[b]{0.33\textwidth}
      \centering
      \subfloat[IEEE 13-bus]{%
        \includegraphics[width=\linewidth]{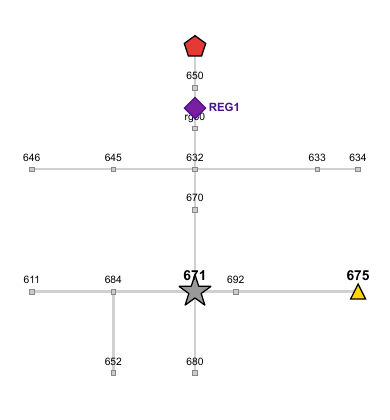}%
        \label{fig:topo-13}%
      }
    \end{minipage}%
    \hfill
    \begin{minipage}[b]{0.65\textwidth}
      \centering
      \subfloat[IEEE 34-bus]{%
        \includegraphics[width=\linewidth]{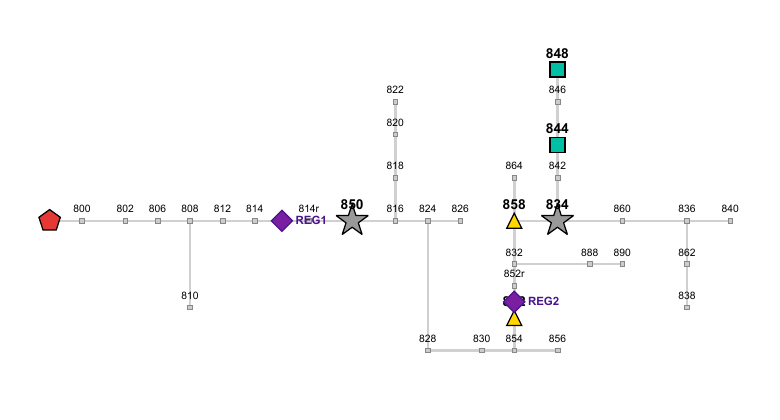}%
        \label{fig:topo-34}%
      }
    \end{minipage}

    \vspace{0.5em}

    \subfloat[IEEE 123-bus]{%
      \includegraphics[width=\linewidth]{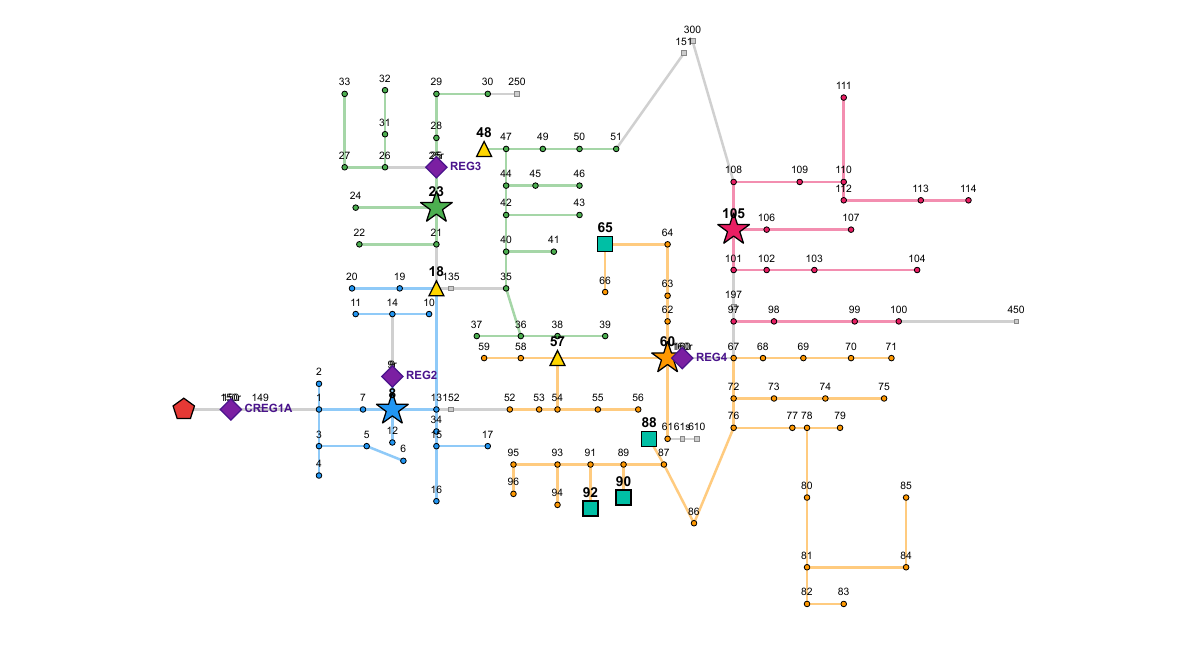}%
      \label{fig:topo-123}%
    }

    \caption{Distribution system topologies used in experiments. Subfigures show the IEEE 13-bus, 34-bus, and 123-bus test feeders with datacenter locations ($\bigstar$), PV systems ($\blacktriangle$), voltage regulators ($\blacklozenge$), and capacitor banks ($\blacksquare$). Zone coloring in (c) indicates the four manually defined control regions of the IEEE 123-bus feeder: buses within the same zone tend to exhibit stronger voltage correlation, but the zones are not physically isolated subnetworks.}
    \label{fig:grid_topologies}
  \end{figure*}

Figure~\ref{fig:grid_topologies} shows the three distribution systems used in our experiments.
For readers unfamiliar with power distribution, a \emph{distribution feeder} is the lower-voltage network that delivers electricity from the bulk power system to end users.
It is connected to the upstream transmission system and centralized generation through a \emph{substation}, shown as the source node in Figure~\ref{fig:grid_topologies}; the substation steps transmission voltage down to feeder voltage and houses the on-load tap changer that adjusts feeder-side voltage in discrete steps.
Given each bus's load and generation, the \emph{power-flow equations} determine the resulting bus voltages and line currents, and OpenG2G solves them via OpenDSS.
Under normal radial operation, power flows from the source bus toward downstream loads.
Because distribution lines have non-negligible impedance, voltage magnitude generally decreases as power flows farther from the source, especially in the absence of intermediate voltage-support devices.
Voltage regulators mitigate this effect by adjusting tap positions to raise or lower downstream voltage, while capacitor banks inject reactive power to support local voltage.
As a first-order rule, increasing the load at a bus tends to reduce nearby and downstream voltages, whereas decreasing the load tends to raise them.
This is the physical mechanism that allows datacenter workload control to provide voltage regulation: by changing GPU power consumption, the datacenter changes the local load seen by the feeder and therefore influences voltage magnitudes.

The .dss files of these systems are adapted from the IEEE test feeder models in the OpenDSS distribution test-case repository~\cite{opendss_ieee_testcases}.
We modify these feeders by adding datacenters, photovoltaic (PV) power generation systems, time-varying load (TVL), and selected capacitor banks to create grid conditions in which workload flexibility can be evaluated for voltage regulation. The IEEE 13-bus feeder contains one datacenter, the IEEE 34-bus feeder contains two datacenters, and the IEEE 123-bus feeder contains four datacenters. 
Including these devices allows OpenG2G to evaluate datacenter-side workload flexibility in realistic distribution settings where datacenters interact with existing feeder topology, PV variability, regulators, and reactive-power compensation. Parameters of DC, PV, and TVL are listed in Table~\ref{tab:system_config}.

For the IEEE 123-bus feeder, we partition the system into four regulator zones following~\cite{li2018decentralized}, and place one datacenter in each zone to study coordination across regions. 
Within each zone, the electrical connections between buses are strong (exhibiting higher voltage correlation), whereas the electrical connections between different zones are relatively weak.
This zoning scheme simplifies the design of the GPU batch size controller.
For example, a PPO controller that directly observes the three-phase voltages of all buses in the IEEE 123-bus feeder would require an observation vector with more than 300 entries.
Such a high-dimensional input would require a larger neural network, more training data, and more interactions with the simulator to learn an effective policy.
Moreover, most buses remain within the voltage limits during most time steps, so feeding every bus voltage into the controller is an inefficient use of the observation space.
Therefore, for the IEEE 123-bus experiments, we use a compact zone-level voltage observation for PPO.
For each of the four zones, the observation includes the minimum voltage, maximum voltage, and integral voltage violation within that zone, resulting in a 12-dimensional voltage feature vector.
This representation preserves the key information needed for control: whether a voltage violation occurs, whether it is an undervoltage or overvoltage event, and which zone is affected.
It therefore allows the PPO controller to respond to system-level voltage problems without requiring full-bus voltage measurements as direct neural-network inputs.

\begin{table}[t]
  \centering
  \caption{Per-feeder system configuration.  GPUs is the total deployable inventory across all DC sites. DC base load is the constant load of ancillary infrastructure such as cooling, accounting for roughly 30\% - 50\% of total DC power consumption, assuming each GPU's maximum power consumption is 700~W \cite{NVIDIAH100}. PV/TVL columns show summed nameplate capacities of photovoltaic generators and time-varying loads. Tap reg.\,is the number of on-load tap-changing regulators in the feeder.}
  \label{tab:system_config}
  \small
  \setlength{\tabcolsep}{4pt}
  \begin{tabular}{l c l c c l l l}
    \toprule
    System & DC sites & DC buses & GPUs & DC base load (kW) & PV (kW) & TVL (kW) & Tap reg. \\
    \midrule
  IEEE\,13   & 1 & 671                    &  7200 &  1500 &  10     & 10     &   1 \\
  IEEE\,34   & 2 & 850, 834               &  2640 &  1650 & 195    & 250    &   2 \\
  IEEE\,123  & 4 & 8, 23, 60, 105         &  4320 &  3024 &  550    & 60     &   4 \\
    \bottomrule
  \end{tabular}
\end{table}

\section{Scenario Library}\label{sec:appendix-ppo-pipeline}

\begin{figure}[t]
  \centering

  \begin{minipage}[t]{0.36\linewidth}\centering
  \subfloat[PV profiles]{\label{fig:dataset_a}\includegraphics[width=\linewidth]{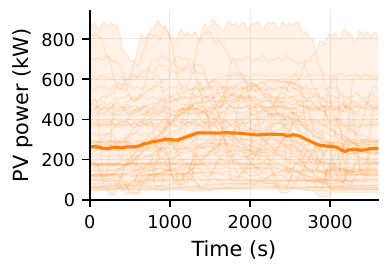}}
  \end{minipage}\hspace{1em}%
  \begin{minipage}[t]{0.36\linewidth}\centering
  \subfloat[TVL profiles]{\label{fig:dataset_b}\includegraphics[width=\linewidth]{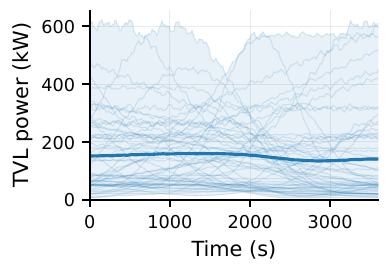}}
  \end{minipage}

  \vspace{0.6em}

  \begin{minipage}[t]{0.36\linewidth}\centering
  \subfloat[Inference GPUs]{\label{fig:dataset_c}\includegraphics[width=\linewidth]{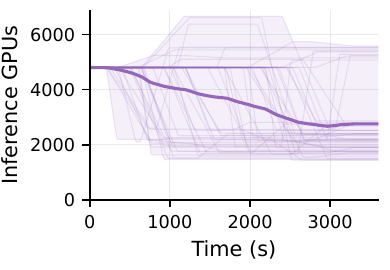}}
  \end{minipage}\hspace{1em}%
  \begin{minipage}[t]{0.36\linewidth}\centering
  \subfloat[Training GPUs]{\label{fig:dataset_d}\includegraphics[width=\linewidth]{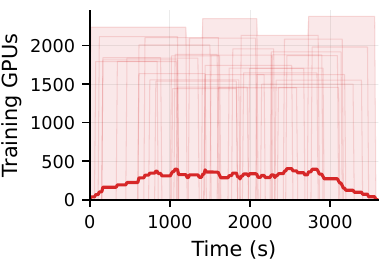}}
  \end{minipage}

  \caption{Diversity of the IEEE-13 training scenario library (235 scenarios). Faint traces show 60 randomly-sampled scenarios; bold
  lines are the across-scenario mean and shaded bands are the min/max envelope. (a) PV power profiles span ${\sim}10{\times}$ in peak.
   (b) Time-varying load profiles vary in both peak and time-of-peak. (c) Inference active GPUs initialize at a fixed deployment (4800 GPUs in total) and change as each scenario's randomly-timed inference ramp adjusts the active replica counts. (d) Training-overlay GPUs:
  69\% of scenarios include a training-workload window with random duration and timing, contributing up to 2400 GPUs at peak.}
  \label{fig:dataset_diversity_ieee13}
  \end{figure}

We construct a per-feeder randomized scenario library to train and evaluate PPO.
Each candidate scenario perturbs the deterministic feeder by sampling: (i) photovoltaic and time-varying-load peak scales together with bus-level profile shapes (rising--falling, flat, midday-dip, etc.), (ii) a coincident GPU training-power overlay, and (iii) one or two inference replica-count ramps per datacenter site, with start times, durations, and amplitudes drawn from independent uniform distributions. The diversity of these perturbs are shown in Figure~\ref{fig:dataset_diversity_ieee13}. 

\begin{figure}[t]
  \centering
  \includegraphics[width=0.2\linewidth]{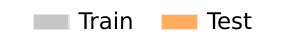}

  \begin{minipage}[t]{0.31\linewidth}\centering
  \subfloat[IEEE-13]{\label{fig:diff_a}\includegraphics[width=\linewidth]{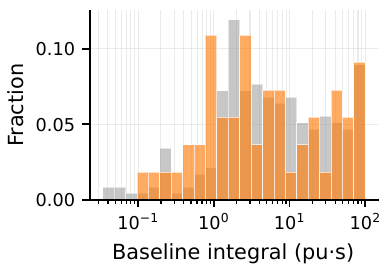}}
  \end{minipage}\hspace{0.5em}%
  \begin{minipage}[t]{0.31\linewidth}\centering
  \subfloat[IEEE-34]{\label{fig:diff_b}\includegraphics[width=\linewidth]{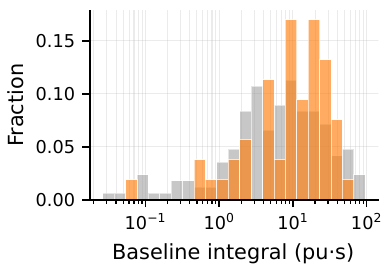}}
  \end{minipage}\hspace{0.5em}%
  \begin{minipage}[t]{0.31\linewidth}\centering
  \subfloat[IEEE-123]{\label{fig:diff_c}\includegraphics[width=\linewidth]{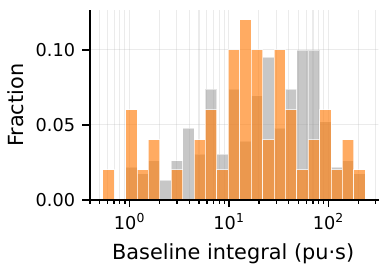}}
  \end{minipage}
  \caption{Per-scenario difficulty distribution (voltage-violation integral under the no-control baseline) for the train and test
  scenario libraries. Log-x axis spans ${\sim}3$ decades on each feeder. Test scenarios cover the same difficulty range as training;
  difficulty scales with feeder size (IEEE-13 median ${\sim}5$ pu$\cdot$s, IEEE-34 ${\sim}6{-}11$ pu$\cdot$s, IEEE-123 ${\sim}17{-}24$ pu$\cdot$s).}
  \label{fig:dataset_difficulty}
  \end{figure}

Each candidate scenario is screened by simulating both no-coordination and OFO references on it: a candidate is admitted only if the no-coordination integral voltage violation exceeds 1 pu$\cdot$s (so trivially-mitigable scenarios are filtered out) and OFO recovers at least $70\%$ of that integral (so infeasible scenarios are filtered out as well).
Accepted scenarios are then split by seed range into disjoint training and test libraries.  Table~\ref{tab:dataset_summary} reports per-feeder library sizes, baseline and OFO integrals, and the mix of under-/over-/both-violation scenarios. Figure~\ref{fig:dataset_difficulty} shows that the training and testing datasets have similar difficulties for controllers in terms of integral voltage violation. 
PPO training (Appendix~\ref{sec:appendix-controller-math}) draws rollouts from the training library, and the test library is used for the head-to-head controller comparison in Section~\ref{sec:controllers}.

\begin{table}[t]
    \centering
    \caption{Scenario library summary. \emph{Baseline} is the mean voltage-violation integral (pu$\cdot$s) under no control; \emph{OFO} is the residual after OFO; \emph{Recov.} is $(\text{Baseline}-\text{OFO})/\text{Baseline}$. The last three columns report the share of accepted scenarios with undervoltage only, overvoltage only, or both violation types present at baseline. Test rows are computed over the $n=50$ scenarios actually used in eval (rather than the full screened library).}
    \label{tab:dataset_summary}
    \small
    \setlength{\tabcolsep}{5pt}
    \begin{tabular}{l l r r r r r r r}
      \toprule
      System & Split & $n$ & Baseline & OFO & Recov. & Under (\%) & Over (\%) & Both (\%) \\
      \midrule
      IEEE-13  & Train & 235 & 22.87 & 1.24 & 94.6\% & 70.3 & 14.9 & 14.9 \\
               & Test  &  50 & 19.52 & 1.04 & 94.7\% & 50.0 & 34.0 & 16.0 \\
      IEEE-34  & Train & 168 & 13.58 & 2.44 & 82.0\% & 33.9 & 39.9 & 26.2 \\
               & Test  &  50 & 14.56 & 2.62 & 82.0\% & 44.0 & 28.0 & 28.0 \\
      IEEE-123 & Train & 231 & 38.08 & 6.36 & 83.3\% & 44.2 & 12.6 & 43.3 \\
               & Test  &  50 & 35.15 & 5.74 & 83.7\% & 44.0 &  6.0 & 50.0 \\
      \bottomrule
    \end{tabular}
  \end{table}

\section{Simulation Settings and Metric Definitions}\label{sec:appendix-setup}
\subsection{Simulation setting}
\begin{figure}[t]
\centering
\includegraphics[width=0.65\linewidth]{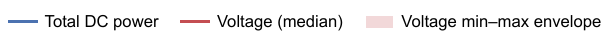}

\begin{minipage}[t]{0.36\linewidth}\centering
\subfloat[Total datacenter power trajectory over a 60-minute horizon]{\label{fig:baseline_ieee13_a}\includegraphics[width=\linewidth]{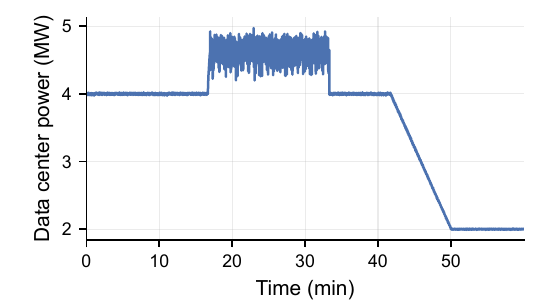}}
\end{minipage}\hspace{1.5em}%
\begin{minipage}[t]{0.36\linewidth}\centering
\subfloat[Voltage envelope across all monitored buses and phases]{\label{fig:baseline_ieee13_b}\includegraphics[width=\linewidth]{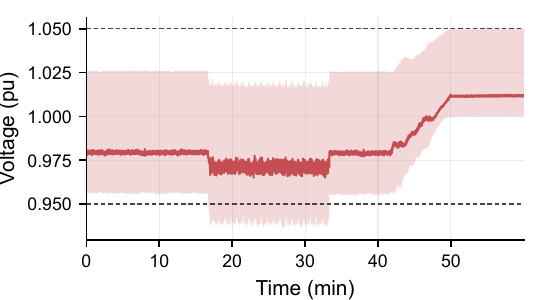}}
\end{minipage}
\caption{Baseline datacenter electricity load on the IEEE 13-bus feeder induces voltage limit violations. Without coordination, the lower limit (0.95 pu) is breached during the training window and the upper limit (1.05 pu) is approached as the load ramps down.}
\label{fig:baseline_ieee13}
\end{figure}

Both Section~\ref{sec:controllers} and Section~\ref{sec:ml-impact} simulate at a base tick of $d_t =0.1$ s and $\bar t= 3,600 s$. Section~\ref{sec:controllers} uses the 50 different scenarios generated in Appendix~\ref{sec:appendix-ppo-pipeline} to evaluate the controllers.
Section~\ref{sec:ml-impact} specializes Section~\ref{sec:controllers}'s IEEE 13-bus, single-datacenter case to one fixed instance: an auxiliary training workload of 2{,}400 GPUs at 400 W/GPU peak runs during $t \in [1000, 2000]$ s, and every model's replica schedule ramps down to 50\% of its initial count during $t \in [2500, 3000]$ s, giving the feeder a realistic mix of steady and transient load. The baseline datacenter power consumption and grid voltage in this scenario are shown in Figure~\ref{fig:baseline_ieee13}.

\subsection{Evaluation Metrics}

We formally define four metrics for evaluating controller performance.
Two of these (integral voltage violation and token throughput) appear in the Section~\ref{sec:controllers} main text; the other two (latency-target violation rate and batch size switching count) characterize inference-service quality and control effort, and are reported per (controller, feeder) pair in Appendix~\ref{sec:appendix-controller-perf}.
These evaluation metrics are not identical to the four terms in the PPO reward function, but they reflect the same four objectives: voltage regulation, token throughput, latency-target adherence, and batch size switching.

\paragraph{Integral voltage violation.}
Let $\bar{t}$ denote the episode duration (run length). The integral voltage violation (pu$\cdot$s) is defined as
\[
  \sum_{t=0}^{\bar{t}} \sum_{b,\phi}\Bigl[\,\max\bigl(0;\,\underline{v} - v_{b,\phi}(t)\bigr)
  \,+\, \max\bigl(0;\,v_{b,\phi}(t) - \overline{v}\bigr)\,\Bigr]\, d_t,
\]
computed over all bus-phase pairs $(b,\phi)$ in the feeder, where $v_{b,\phi}(t)$ is the voltage at bus $b$ on phase $\phi$, and $\underline{v}=0.95$~pu, $\overline{v}=1.05$~pu are the allowable bounds, and $d_t$ is the simulation time resolution. The metric captures both duration and severity of excursions: a $0.10$~pu violation for 10~s equals a $0.01$~pu violation for 100~s.

\paragraph{Token throughput.}
The mean throughput (tok/s) is defined as
\[
   \frac{1}{\bar{t}\,}\,\sum_{t=0}^{\bar{t}} \sum_{i \in\mathcal{I}} T_i(t)\, d_t,
\]
where $T_i(t)$ is the instantaneous token-generation rate of LLM $i$ at time $t$ and the sum runs over all LLMs deployed across all datacenters.

\paragraph{Latency-target violation rate.}
Let $N$ be the number of control steps in the episode and $\mathcal{I}$ the set of deployed LLMs across all datacenters. The latency-target violation rate is defined as
\[
  \frac{1}{\bar{t}\,|\mathcal{I}|}\,\sum_{t=0}^{\bar{t}}\sum_{i\in\mathcal{I}}
  \mathbf{1}\!\left[\,\mathrm{ITL}_i(t) > \overline{\mathrm{ITL}}_i\,\right],
\]
where $\mathrm{ITL}_i(t)$ is the observed inter-token latency of LLM $i$ at step $t$, $\overline{\mathrm{ITL}}_i$ is its per-model latency target, and $\mathbf{1}[\cdot]$ is the indicator function.

\paragraph{Batch size switching.}
The batch size switching count is defined as
\[
  \sum_{t=0}^{\bar{t}-1} \sum_{i\in\mathcal{I}} \mathbf{1}\!\left[\, b_i(t+1) \,\ne\, b_i(t)\,\right],
\]
where $b_i(t)\in\mathcal{B}_i$ is the realized (discrete) batch size assigned to LLM $i$ at control step $t$.

\section{Additional Controller Studies}\label{sec:appendix-controller-perf}
\subsection{Full evaluation results of controllers}

Table~\ref{tab:full_controller_eval} complements the main voltage and throughput results by reporting control effort and latency adherence.
All three active controllers introduce batch size switching because they regulate voltage by changing GPU power consumption, but the switching frequency remains moderate.
Across the one-hour simulation window, each deployed model changes its batch size only about four to five times on average.
The total switching count generally increases with feeder complexity and the number of datacenters, reflecting the greater coordination burden in larger systems where multiple datacenters at different buses must respond to spatially heterogeneous voltage problems.
PPO also exhibits larger variability in batch switching on the IEEE 123-bus feeder ($30.2\pm9.7$), indicating that different random seeds can learn noticeably different coordination patterns.

The latency-target violation rate remains relatively small for all controllers, but the trends reveal different service-quality trade-offs.
OFO keeps latency violations near $2\%$ across all three feeders, suggesting that its explicit latency-aware objective helps maintain service quality while providing voltage regulation.
PPO has higher latency violation rates, increasing from $3.0\%$ on IEEE 13 to $5.5\%$ on IEEE 123, which is consistent with its tendency to preserve higher throughput and learn more aggressive policies in some seeds.
Droop control has low latency violations because it often reduces batch sizes conservatively, but this also limits its throughput and voltage-regulation performance in larger systems.
Overall, the table shows that controller evaluation in OpenG2G must consider not only voltage violation and throughput, but also the operational costs of frequent batch changes and the service-quality impact of latency violations.

 \begin{table}[t]
      \centering
      \caption{Controller evaluation with four metrics over 50 scenarios. Each (controller, feeder) pair is run five times. PPO entries report mean $\pm$ across-seed standard deviation (best checkpoint per training seed); droop and OFO produced identical values across runs and are reported as single numbers.}
      \label{tab:full_controller_eval}
      \small
      \begin{tabular}{l l l l l l}
        \toprule
        System & Controller & Voltage.\ viol (pu$\cdot$s) & Tput (M tok/s) & Batch $\Delta$ & ITL\ viol(\%) \\
        \midrule
      IEEE\,13   & No Coordination &  19.52         &  5.25         &   0.0         &   0.0         \\
                 & Droop           &   1.39         &  4.11         &  14.9         &   1.8         \\
                 & PPO             & $1.62\pm0.44$  & $4.43\pm0.91$ & $22.5\pm5.5$  & $3.0\pm1.4$   \\
                 & OFO             &   1.04         &  4.17         &  18.5         &   1.9         \\
      IEEE\,34   & No Coordination &  14.56         &  1.30         &   0.0         &   0.0         \\
                 & Droop           &   7.88         &  1.11         &  20.5         &   2.5         \\
                 & PPO             & $5.81\pm1.07$  & $1.35\pm0.15$ & $17.7\pm1.8$  & $3.9\pm0.9$   \\
                 & OFO             &   2.62         &  1.24         &  21.8         &   2.4         \\
      IEEE\,123  & No Coordination &  35.15         &  1.46         &   0.0         &   0.0         \\
                 & Droop           &  29.70         &  1.20         &  21.4         &   0.0         \\
                 & PPO             & $10.64\pm3.83$ & $1.51\pm0.44$ & $30.2\pm9.7$  & $5.5\pm2.1$   \\
                 & OFO             &   5.74         &  1.60         &  30.8         &   2.3         \\
        \bottomrule
      \end{tabular}
    \end{table}

\subsection{PPO training dynamics}
\begin{figure}[t]
  \centering
  \includegraphics[width=0.45\linewidth]{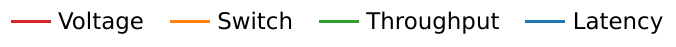}
  
  \begin{minipage}[t]{0.31\linewidth}\centering
  \subfloat[IEEE-13]{\label{fig:rewbreak_a}\includegraphics[width=\linewidth]{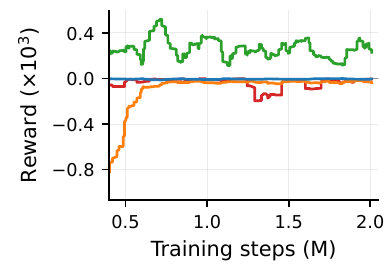}}
  \end{minipage}\hspace{0.5em}%
  \begin{minipage}[t]{0.31\linewidth}\centering
  \subfloat[IEEE-34]{\label{fig:rewbreak_b}\includegraphics[width=\linewidth]{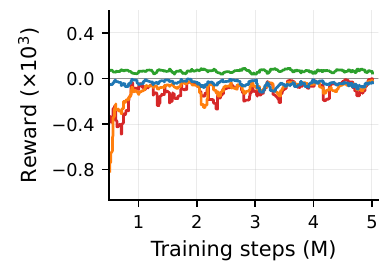}}
  \end{minipage}\hspace{0.5em}%
  \begin{minipage}[t]{0.31\linewidth}\centering
  \subfloat[IEEE-123]{\label{fig:rewbreak_c}\includegraphics[width=\linewidth]{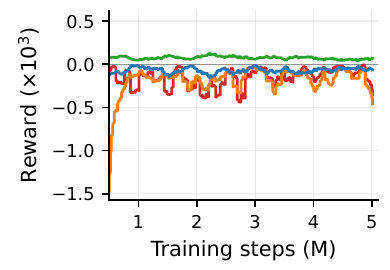}}
  \end{minipage}

  \caption{Per-episode reward breakdown during PPO training (rolling mean over 50 episodes) for one representative run per feeder. }
  \label{fig:training_reward_breakdown}
\end{figure}

Figure~\ref{fig:training_reward_breakdown} decomposes the PPO reward during training into voltage, switching, throughput, and latency components.
Across all three feeders, the switching penalty is the dominant negative term at the beginning of training, indicating that the randomly initialized policy frequently changes batch sizes and creates unstable control actions.
As training progresses, this penalty quickly decreases, showing that PPO first learns to suppress unnecessary batch oscillations.
After this initial stabilization phase, the remaining reward components reveal the controller's operating trade-off.
The throughput term stays positive because the agent is rewarded for maintaining productive inference service, while the voltage and latency terms become active when aggressive batch size choices create grid or service-quality violations.
The larger IEEE 34- and 123-bus systems exhibit more persistent voltage and switching penalties than the IEEE 13-bus case, reflecting the increased difficulty of coordinating multiple datacenters across a more complex feeder.
Thus, this reward breakdown provides a diagnostic view of PPO training: the agent first learns stable actions, and then negotiates the multi-objective trade-off among voltage regulation, throughput, latency, and control effort.

\subsection{Controller parameter sweep} 
Table~\ref{tab:param-sweep} reports a sensitivity study for the most influential tunable parameter of each controller: the throughput weight $\alpha_T$ in OFO and the voltage-cost weight $w_V$ in PPO.
These two parameters are selected because they directly control the trade-off between inference throughput and voltage-regulation performance shown in Figure~\ref{fig:tradeoff}.

For OFO, the voltage-regulation weight is implicitly determined by the dual variables associated with the voltage constraints, so the main adjustable objective weight is $\alpha_T$.
Increasing $\alpha_T$ encourages larger batch sizes and higher throughput, but can weaken voltage regulation; decreasing it improves voltage performance at the cost of throughput.
The cases $\alpha_T=10^{-4}$ and $\alpha_T=10^{-5}$ achieve the same mean integral violation of $1.04$ pu$\cdot$s, but $\alpha_T=10^{-4}$ provides higher throughput, so we use $\alpha_T=10^{-4}$ in the main experiments. 

For PPO, $w_V$ determines how strongly voltage violations are penalized during training.
A larger $w_V$ guides the policy toward better voltage regulation, improving the mean integral violation from $2.46$ pu$\cdot$s at $w_V=1$k to $1.28$ pu$\cdot$s at $w_V=5$k.
However, further increasing the weight to $10$k degrades held-out performance, suggesting that an overly large voltage penalty can lead to a less generalizable policy.
Therefore, we use $w_V=5$k as the default PPO setting.

\begin{figure}[t]
\centering
\includegraphics[width=0.45\linewidth]{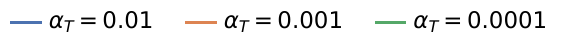}

\begin{minipage}[t]{0.36\linewidth}\centering
\subfloat[Batch size (Qwen3 30B A3B)]{\label{fig:w_throughput_a}\includegraphics[width=\linewidth]{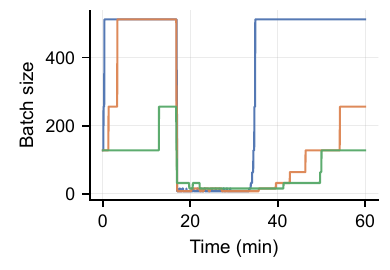}}
\end{minipage}\hspace{1.5em}%
\begin{minipage}[t]{0.36\linewidth}\centering
\subfloat[Voltage envelope]{\label{fig:w_throughput_b}\includegraphics[width=\linewidth]{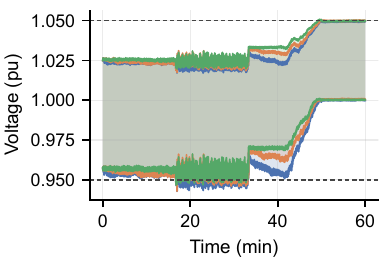}}
\end{minipage}
\caption{Effect of throughput weight $\alpha_{T}$ on OFO control for IEEE-13. Higher $\alpha_{T}$ drives batch sizes toward the maximum in (a), increasing throughput but widening voltage violations below the 0.95\,pu limit in (b). Reducing $\alpha_{T}$ from 0.01 to 0.0001 cuts integral violation by 21$\times$ at the cost of 44\% lower average throughput.}
\label{fig:w_throughput_sweep}
\end{figure}

Figure~\ref{fig:w_throughput_sweep} further illustrates how the OFO throughput weight $\alpha_T$ shapes the controller behavior.
A larger $\alpha_T$ places more emphasis on inference throughput, so the controller keeps the batch size high whenever the resulting voltage profile remains close to the allowable range.
This leads to higher throughput but leaves less margin for voltage regulation, as shown by the wider voltage excursions near the lower voltage limit.
In contrast, a smaller $\alpha_T$ makes the controller more conservative: it reduces the batch size earlier and maintains lower batch sizes during the undervoltage period, thereby reducing integral voltage violation at the cost of throughput.
This confirms that $\alpha_T$ directly controls the throughput--voltage-regulation trade-off in OFO.

\begin{figure}[t]
  \centering
  \includegraphics[width=0.65\linewidth]{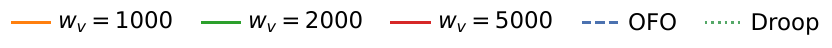}

  \begin{minipage}[t]{0.36\linewidth}\centering
  \subfloat[Training-time voltage cost]{\label{fig:wv_sweep_a}\includegraphics[width=\linewidth]{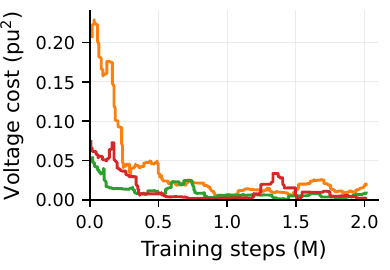}}
  \end{minipage}\hspace{1.5em}%
  \begin{minipage}[t]{0.36\linewidth}\centering
  \subfloat[Eval-time integral per checkpoint]{\label{fig:wv_sweep_b}\includegraphics[width=\linewidth]{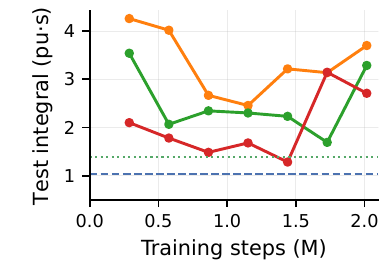}}
  \end{minipage}
  \caption{Effect of voltage weight $w_v$ on PPO training and evaluation for IEEE-13. (a) Per-episode squared voltage cost
  $\sum_{t}\mathcal{P}_V(t)$ during training (rolling mean over 100 episodes, with $w_v$ divided out for
  cross-variant comparability). (b) Mean integral voltage violation on the 50-scenario test set, evaluated at each checkpoint; OFO and droop baselines shown as horizontal references. All variants share identical hyperparameters except $w_v$. Increasing $w_v$ from 1000 to 5000 cuts test-time integral violation from 2.46 to 1.28~pu$\cdot$s ($-$48\%) while changing throughput by less than 2\%.}
  \label{fig:w_voltage_sweep}
\end{figure}

Figure~\ref{fig:w_voltage_sweep} shows how the PPO voltage-cost weight $w_V$ affects both training dynamics and held-out evaluation performance.
During training, larger $w_V$ encourages the policy to reduce voltage cost more aggressively, leading to a faster decrease in the per-episode voltage-cost signal.
For all tested values of $w_V$, the training voltage cost reaches a low level after roughly $1$M steps and then remains relatively stable.
However, the evaluation results show that the checkpoint with the lowest integral voltage violation is not necessarily the final checkpoint.
Instead, the best test performance occurs at an intermediate checkpoint, around $1.5$M training steps for the best-performing setting.
This gap between training convergence and held-out performance suggests that longer training does not always improve generalization, and supports our use of checkpoint selection based on the 50-scenario evaluation set.

 \begin{table}[t]
    \centering
    \small
    \caption{Effect of the throughput weight $\alpha_T$ (OFO) and the voltage-cost weight $w_V$ (PPO) on IEEE-13
  controller performance. \emph{Best ckpt} reports the checkpoint minimizing mean integral on the 50-scenario test set;
  reported values are mean $\pm$ std over those scenarios. The droop controller is shown as a reference.}
    \label{tab:param-sweep}
    \begin{tabular}{llllll}
      \toprule
      Controller & Best ckpt & Voltage viol.(pu$\cdot$s) & Tput (M tok/s) & Batch $\Delta$ & ITL 
  viol.(\%)\\
      \midrule
      No Coordination                & --        & $19.52 \pm 27.84$ & $5.25 \pm 1.63$  & $0$              & $0.0 \pm 0.0$
  \\
      Droop control              & --        & $1.39 \pm 2.29$   & $4.11 \pm 2.04$  & $14.9 \pm 9.1$   & $1.8 \pm 2.8$
  \\
      \midrule
      OFO ($\alpha_T{=}10^{-2}$) & --        & $1.23 \pm 2.06$   & $10.40 \pm 2.73$ & $112.0 \pm 80.2$ & $5.5 \pm 3.6$
  \\
      OFO ($\alpha_T{=}10^{-3}$) & --        & $1.07 \pm 2.06$   & $7.74 \pm 3.07$  & $44.8 \pm 28.2$  & $2.3 \pm 2.7$
  \\
      OFO ($\alpha_T{=}10^{-4}$) & --        & $\mathbf{1.04 \pm 2.06}$ & $\mathbf{4.61 \pm 2.42}$ & $\mathbf{26.5 \pm
  33.7}$ & $\mathbf{1.9 \pm 2.5}$ \\
      OFO ($\alpha_T{=}10^{-5}$) & --        & $1.04 \pm 2.06$   & $4.17 \pm 2.21$  & $18.5 \pm 10.5$  & $1.9 \pm 2.4$
  \\
      \midrule
      PPO ($w_V{=}1$k)           & $1.15$\,M & $2.46 \pm 3.63$   & $5.49 \pm 3.47$  & $18.2 \pm 7.6$   & $7.4 \pm 6.5$
  \\
      PPO ($w_V{=}2$k)           & $1.73$\,M & $1.69 \pm 2.77$   & $4.89 \pm 3.48$  & $20.1 \pm 7.8$   & $7.9 \pm 5.5$
  \\
      PPO ($w_V{=}5$k)           & $1.44$\,M & $\mathbf{1.28 \pm 2.36}$ & $\mathbf{5.38 \pm 3.33}$ & $\mathbf{29.8 \pm
  9.1}$ & $\mathbf{4.5 \pm 4.3}$ \\
      PPO ($w_V{=}10$k)          & $2.02$\,M & $1.65 \pm 2.65$   & $4.69 \pm 3.11$  & $21.8 \pm 7.5$   & $3.9 \pm 4.4$
  \\
      \bottomrule
    \end{tabular}
  \end{table}

\section{Model Size on H100}\label{sec:appendix-h100-model-size}

\begin{figure}[t]
\centering
\includegraphics[width=\linewidth]{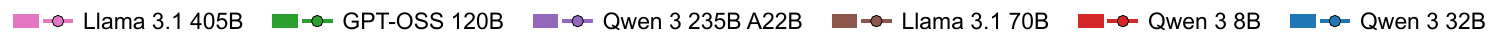}

\begin{minipage}[t]{0.36\linewidth}\centering
\subfloat[Integral voltage violation]{\label{fig:model-size-h100-violation}\includegraphics[width=\linewidth]{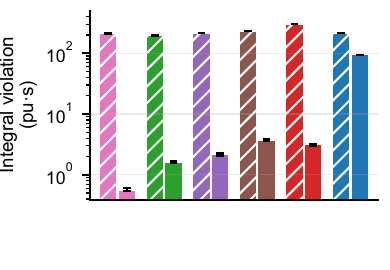}}
\end{minipage}\hspace{1.5em}%
\begin{minipage}[t]{0.36\linewidth}\centering
\subfloat[Power--throughput curve]{\label{fig:model-size-h100-pareto}\includegraphics[width=\linewidth]{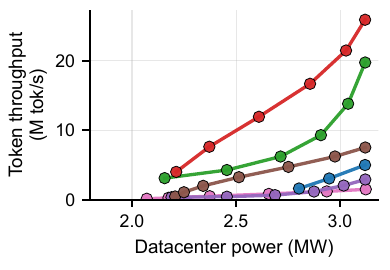}}
\end{minipage}
\caption{Model size and architecture on H100: six models, ordered left-to-right by decreasing datacenter power range. In (a), hatched bars are without coordination; solid bars use OFO~\cite{g2g-powerup26}. Models with wider span in (b) show deeper reductions in (a).}
\label{fig:model-size-h100}
\end{figure}

We complement Section~\ref{sec:ml-impact-model-design}'s B200 model-size study with the same comparison on H100, holding the match-peak anchor fixed (3.12 MW at Qwen 3 8B, 4{,}800 replicas on H100).
Figure~\ref{fig:model-size-h100} compares six models, all served on H100 GPUs: four B200-counterpart models from Figure~\ref{fig:model-size-b200} (Qwen 3 8B, Qwen 3 32B, GPT-OSS 120B, Qwen 3 235B A22B Instruct), plus Llama 3.1 70B and Llama 3.1 405B (for which v3 does not yet have B200 measurements).
Qwen 3 30B A3B Instruct is absent from the H100 ladder for the same fit-artifact reason Qwen 3 14B is absent from the B200 ladder: the logistic-mean ITL at the 50 ms target sits well under the deadline, but the underlying two-component-lognormal ITL mixture has a heavy tail that produces excess misses under OFO.
Each remaining model is match-peak sized to the shared 3.12 MW anchor at the largest batch size meeting its ITL target (50 ms for chat, 100 ms for Qwen 3 235B A22B and Llama 3.1 70B, 120 ms for Llama 3.1 405B).
The flexibility ordering tracks the B200 ladder qualitatively: the fractional per-replica power range across the feasible batch range, the same mechanism identified in Section~\ref{sec:ml-impact-model-design}, determines datacenter power range and the residual under coordination.
Qwen 3 32B at 1 GPU on H100 illustrates the opposite extreme of the hardware-memory mechanism from Section~\ref{sec:ml-impact-deployment}: the $80$ GB cap on H100 leaves only a 3-batch ladder, giving a $0.32$ MW datacenter power range and an OFO residual of $93.9$ pu$\cdot$s; coordination has essentially no room to move.
Llama 3.1 405B at a 120 ms target carries a $\sim 4\%$ ITL miss rate under OFO at match-peak, reflecting the same ITL-tail artifact as the dropped models above.

\section{Additional Precision Pairs}\label{sec:appendix-precision}

The main text presents the BF16 vs FP8 comparison on Qwen 3 235B A22B Instruct at H100 $\times$ 8 GPU on chat.
Two further matched-hardware pairs in the Qwen 3 235B A22B family share the same mechanism: Thinking on H100 $\times$ 8 GPU and on B200 $\times$ 4 GPU, both at a 100 ms reasoning target (Table~\ref{tab:appendix-precision-pairs}).
Match-peak sizing to the same 3.12 MW anchor gives 906 and 991 replicas, respectively.

\begin{table}[t]
\centering
\small
\subfloat[Qwen 3 235B A22B Thinking, H100 $\times$ 8 GPU, 100 ms]{%
\begin{tabular}{l r r r r r}
\toprule
Precision & $B_{\max}$ & Range (MW) & No coord.\ viol. & OFO viol. & OFO thpt (M tok/s) \\
\midrule
FP8  & 128 & 0.24 & $484 \pm 2$ & $314 \pm 1$  & 0.25 \\
BF16 & 32  & 0.55 & $197 \pm 1$ & $43 \pm 0.5$ & 0.36 \\
\bottomrule
\end{tabular}}\\[0.6em]
\subfloat[Qwen 3 235B A22B Thinking, B200 $\times$ 4 GPU, 100 ms]{%
\begin{tabular}{l r r r r r}
\toprule
Precision & $B_{\max}$ & Range (MW) & No coord.\ viol. & OFO viol. & OFO thpt (M tok/s) \\
\midrule
FP8  & 512 & 0.27 & $341 \pm 3$ & $132 \pm 2$ & 0.29 \\
BF16 & 256 & 0.60 & $265 \pm 3$ & $27 \pm 1$  & 0.40 \\
\bottomrule
\end{tabular}}
\caption{Two additional BF16 vs FP8 pairs in the Qwen 3 235B A22B~\cite{qwen3-arxiv25} family. As in the main text, BF16 has a wider datacenter power range and lower OFO integral voltage violation; FP8 reaches higher OFO throughput.}
\label{tab:appendix-precision-pairs}
\end{table}

The BF16/FP8 datacenter-power-range ratio falls between $2.2\times$ and $2.8\times$ across all three pairs.
The OFO violation ratio is noisier: it compounds the power-range gap with a nonlinear controller response, so when BF16 residuals land near the simulator's noise floor the ratio reads as $40\times$ (the main-text pair), while at residuals well above noise (the two pairs above) it falls back toward the power-range ratio.

\paragraph{Why each additional pair differs from the main-text pair.}
The Thinking-on-H100 pair differs on task and model variant (chat vs.\ reasoning, Instruct vs.\ Thinking).
v3 measured Thinking only up to batch size 32 on BF16 at H100 $\times$ 8 GPU (a v3-coverage limit, not a latency limit at 140 ms), so the BF16 variant has a narrow feasible batch range even though the latency target is looser.
Match-peak sizing needs 906 replicas to reach 3.12 MW; the higher replica count raises the baseline stress, and both residuals land well above noise.
The Thinking-on-B200 pair differs on hardware and parallelism (H100 $\times$ 8 GPU vs.\ B200 $\times$ 4 GPU).
Per-replica compute halves but per-GPU compute roughly doubles, so per-replica peak power is similar to the main-text pair; match-peak gives 991 replicas.
BF16 $B_{\max} = 256$ and FP8 $B_{\max} = 512$ return the batch range to roughly the main-text scale, and residuals are again well above noise.

\section{Parallelism on H100}\label{sec:appendix-h100-parallelism}

\begin{figure}[t]
\centering
\begin{minipage}[t]{0.36\linewidth}\centering
\includegraphics[width=0.6\linewidth]{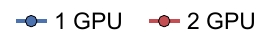}\\[-2pt]
\subfloat[Qwen 3 30B A3B Instruct, 1 vs 2 GPU]{\label{fig:parallelism-h100-pareto}\includegraphics[width=\linewidth]{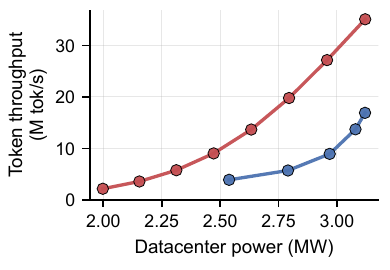}}
\end{minipage}
\caption{Parallelism on H100: Qwen 3 30B A3B Instruct~\cite{qwen3-arxiv25} at 1 vs 2 GPU, match-peak sized.}
\label{fig:parallelism-h100}
\end{figure}

We complement Section~\ref{sec:ml-impact-deployment}'s B200 parallelism study with one additional pair on H100, holding the match-peak anchor fixed.
Figure~\ref{fig:parallelism-h100} compares Qwen 3 30B A3B Instruct at 1 GPU vs 2 GPU on lm-arena-chat at a 100 ms target, both match-peak sized.
The deadline is relaxed from the 50 ms chat default because the 1-GPU fit's ITL tail produces excess misses at 50 ms even though the mean stays well under the target.
Doubling tensor parallelism widens the datacenter power range from $0.58$ to $1.12$ MW and drops OFO's residual integral violation from $35.5$ to $0.25$ pu$\cdot$s, the same direction as the main-text GPT-OSS 120B pair.
Mechanism: each GPU moves further from saturation at small batch sizes, lowering the floor of the per-GPU curve more than the ceiling.
The 1-GPU configuration additionally carries a $23\%$ ITL miss rate under OFO (the ITL-tail artifact persists even at 100 ms); the 2-GPU configuration has no ITL misses.

\section{Reproducibility}\label{sec:appendix-repro}

\paragraph{Code availability.}
OpenG2G is released under an Apache-2.0 license.
The ML.ENERGY Benchmark v3 dataset (consumed by OpenG2G's datacenter backend) is also Apache-2.0.
A \texttt{pip install} command installs the library; all experiments in this paper are driven by scripts in the public repository: \url{https://github.com/gpu2grid/openg2g}.
The example scripts referenced below require the \texttt{examples} dependency group:
\begin{minted}{bash}
uv sync --group examples   # or: pip install --group examples
\end{minted}
Complete documentation, including environment preparation and setup, is available inside the \texttt{docs/} directory of the repository.

\paragraph{Controller experiments (Section~\ref{sec:controllers}).}
The pipeline has three steps. Commands below are for the IEEE 13-bus feeder; replace \texttt{ieee13} with \texttt{ieee34} or \texttt{ieee123} for the other two feeders, adjusting PPO training hyperparameters per Table~\ref{tab:ppo_hyperparams} and the PPO observation mode per Table~\ref{tab:obs_modes}.

\textit{Step 1: Build training and test scenario libraries.}
\begin{minted}{bash}
python examples/offline/build_scenario_library.py \
    --system ieee13 \
    --n-candidates 500 --seed-start 0 \
    --tag train_n500

python examples/offline/build_scenario_library.py \
    --system ieee13 \
    --n-candidates 150 --seed-start 1000 \
    --tag test_n150
\end{minted}

\textit{Step 2: Train PPO (one invocation per random seed).}
\begin{minted}{bash}
python examples/offline/train_ppo.py \
    --system ieee13 \
    --total-timesteps 2000000 \
    --total-duration-s 3600 \
    --n-steps 3600 \
    --hidden-dims 128 128 128 \
    --learning-rate 1e-4 \
    --ent-coef 0.01 \
    --action-mode delta \
    --w-voltage 5000 --w-throughput 0.05 --w-latency 0.01 --w-switch 0.5 \
    --n-envs 8 --seed 1 \
    --scenario-library outputs/ieee13/scenario_library/train_n500/library.pkl \
    --no-ofo-baseline --truncate-episode \
    --output-dir ppo
\end{minted}

\textit{Step 3: Evaluate all four controllers (No Coordination, droop, PPO, OFO).}
\begin{minted}{bash}

python examples/offline/evaluate_controllers.py \
    --system ieee13 \
    --ppo-models examples/offline/outputs/ieee13/ppo/ppo_model.zip
    --scenario-library outputs/ieee13/scenario_library/test_n150/library.pkl \
    --n-scenarios 50 \
    --obs-mode full-voltage \
    --include-rule-based \
    --use-display-names \
    --output-dir eval_controllers \
    --log-level INFO
\end{minted}

\paragraph{AI model and deployment studies (Section~\ref{sec:ml-impact}, Appendices~\ref{sec:appendix-h100-model-size}, \ref{sec:appendix-precision}, and~\ref{sec:appendix-h100-parallelism}).}
Run the four standalone drivers from the repository root:
\begin{minted}{bash}
python examples/model_insights/model_size.py
python examples/model_insights/precision.py
python examples/model_insights/parallelism.py
python examples/model_insights/hardware.py
\end{minted}

\paragraph{Compute resources.}
All experiments run on CPU; no GPU compute is required.
The Section~\ref{sec:ml-impact} studies of AI model and deployment choices take roughly 30 minutes total on an M5 Max MacBook Pro with 64 GB of memory; individual experiments take 1--2 minutes.
The Section~\ref{sec:controllers} controller experiments run on a university Slurm cluster, with each task allocated 10 CPUs and 64 GB of memory.
PPO training takes 12--24 hours per model, and the main text reports results from 15 trained PPO models (5 random seeds $\times$ 3 feeders); the Appendix shows an additional 3 PPO models with different weight parameters for sensitivity analysis.
After training, evaluating all controllers across each feeder takes about 1 hour, also on CPU.\@
Storage requirements are dominated by the 99 GB ML.ENERGY Benchmark dataset download; all other artifacts are negligible.
The full research project required additional compute beyond what is reported here, including preliminary experiments and abandoned configurations.

\section{Limitations}\label{sec:appendix-limitations}

OpenG2G is a \emph{simulation} substrate: results depend on the fidelity of the underlying benchmark (per-configuration power and latency distributions) and the grid simulator (power-flow model, network topology).
Our power-curve substrate is measured, not derived from first principles; models not in the benchmark require additional measurement before they can be simulated.
Grid-side, our default OpenDSS-based backend captures quasi-steady-state voltages; faster transient phenomena (sub-cycle events, inrush currents) are out of scope unless a different grid backend is plugged in.
The default datacenter backend models LLM inference; training appears as an aggregate power overlay rather than a controllable workload.
The default grid backend models distribution feeders; the controller interface accommodates other grid types such as transmission, but our experiments do not yet validate them.
Each PPO controller is trained per-feeder, so cross-feeder generalization is not studied.
Simulation runtime scales with feeder size, the number of inference replicas being simulated, and the number of scenarios, but remains tractable on commodity CPUs (Appendix~\ref{sec:appendix-repro}).

Despite these limitations, we believe OpenG2G's existing contributions are significant, and nothing in OpenG2G's design precludes future extensions to address the aforementioned limitations.
In fact, many of these limitations are planned to be addressed soon or currently being addressed by the authors, and we look forward to sharing those extensions in our open-source library.

\section{Broader Impacts}\label{sec:appendix-impacts}

Coordinating AI datacenters with their host grids offers a path to relaxing the grid-side constraint that currently bottlenecks AI infrastructure deployment.
If effective coordination becomes commonplace, datacenter operators may be able to site new capacity on feeders that would otherwise be oversubscribed, and grid operators may gain a new class of controllable demand-response resource.
This may improve the economics of AI infrastructure deployment and accelerate the pace of AI innovation, while also helping stabilize availability, reliability, and price for all grid customers.
On the risk side, poorly designed coordination (for example, a controller that prioritizes grid stability too aggressively over service-level objectives) could shift costs from grid operators onto end users of AI services in the form of higher latency or throughput throttling.
OpenG2G is intended to support rigorous comparison of coordination methods before such methods are deployed, and help to mitigate this risk by surfacing the service-quality trade-offs of different approaches.

\end{document}